\documentclass[runningheads]{llncs}

\usepackage{eccv}

\usepackage{eccvabbrv}

\usepackage{graphicx}
\usepackage{booktabs}

\usepackage[accsupp]{axessibility}  % Improves PDF readability for those with disabilities.

\usepackage{booktabs}
\usepackage{multirow} 
\usepackage{tabularx} 
\usepackage{array}
\usepackage{amsmath,amssymb}
\usepackage{booktabs}
\usepackage{algorithm}
\usepackage{algorithmic}
\usepackage{threeparttable}
\usepackage{colortbl,xcolor}
\usepackage{siunitx}
\usepackage{marvosym}
\usepackage{pifont}
\usepackage{makecell}
\newcommand{\cmark}{\ding{51}} % ✓
\newcommand{\xmark}{\ding{55}} % ✗
\usepackage{adjustbox}
\usepackage{xcolor}
\usepackage{colortbl}
\usepackage{enumitem}
\definecolor{bestbg}{RGB}{235,245,255}      % light blue
\definecolor{secondbg}{RGB}{245,245,245}    % light gray
\definecolor{groupbg}{RGB}{230,230,230}     % very light gray for group rows

\newcommand{\Best}[1]{\textbf{#1}}
\newcommand{\Second}[1]{\underline{#1}}
\definecolor{ModifyBlue}{RGB}{0,102,204}
\definecolor{UnsureGreen}{RGB}{0,153,0}

\usepackage{hyperref}

\usepackage{orcidlink}

\begin{document}

\title{Towards Temporal Compositional Reasoning in Long-Form Sports Videos} 

\titlerunning{Towards Temporal Compositional Reasoning in Long-Form Sports Videos}

\author{Siyu Cao\inst{1,2}\orcidlink{0009-0002-3024-4500} \and
Lu Zhang{\textsuperscript{\textrm{\Letter}}}\inst{1,2}\orcidlink{0000-0001-6240-5300} \and
Ruizhe Zeng\inst{1,2}\orcidlink{0009-0004-3513-2297} \and
Zhi-yong Liu{\textsuperscript{\textrm{\Letter}}}\inst{1,2,3}\orcidlink{0000-0003-2148-1846}
}

\authorrunning{Cao et al.}

\institute{MAIS, Institute of Automation, Chinese Academy of Sciences,\\
Beijing 100190, China\\
\email{\{caosiyu2024, lu.zhang, zengruizhe2022, zhiyong.liu\}@ia.ac.cn}\and
School of Artificial Intelligence, University of Chinese Academy of Sciences,\\
Beijing 100049, China \and
Nanjing Artificial Intelligence Research of IA, Nanjing, 211100, China}
\maketitle

\begin{abstract}
Sports video analysis is a challenging domain for multimodal understanding because it involves complex and dynamic human activities.
Despite rapid progress in Multimodal Large Language Models (MLLMs), long-horizon reasoning in sports videos remains difficult, as answering questions requires both locating and integrating temporally sparse evidence into reasoning.
We attribute this limitation to two closely related factors: insufficient supervision over temporally dispersed evidence and the lack of methods for explicit temporal evidence localization and justification.
To address these gaps, we introduce \textbf{SportsTime}, a large-scale benchmark for long-form sports video understanding, comprising 14K+ open-ended QA pairs and 50K+ step-wise temporal evidence annotations.
Building on SportsTime, we propose \textbf{Chain-of-Time Reasoning (CoTR)}, which treats reasoning as a process of temporally grounded evidence composition. Specifically, during training, CoTR introduces a temporal-reward GRPO to encourage temporally grounded reasoning.
During inference, it employs an anchor-observe-infer evidence-seeking loop to iteratively localize, verify, and compose temporal evidence before producing the final answer.
Experiments show that SportsTime exposes substantial gaps in current MLLMs, while CoTR yields consistent gains over strong baselines, improving both temporal compositional reasoning performance and step-wise grounding quality. The dataset and code are available at \url{https://github.com/ustiniansy/SportsTime}.
\end{abstract}

\begin{figure}[h]
  \centering
  \resizebox{\linewidth}{0.34\textheight}{\includegraphics{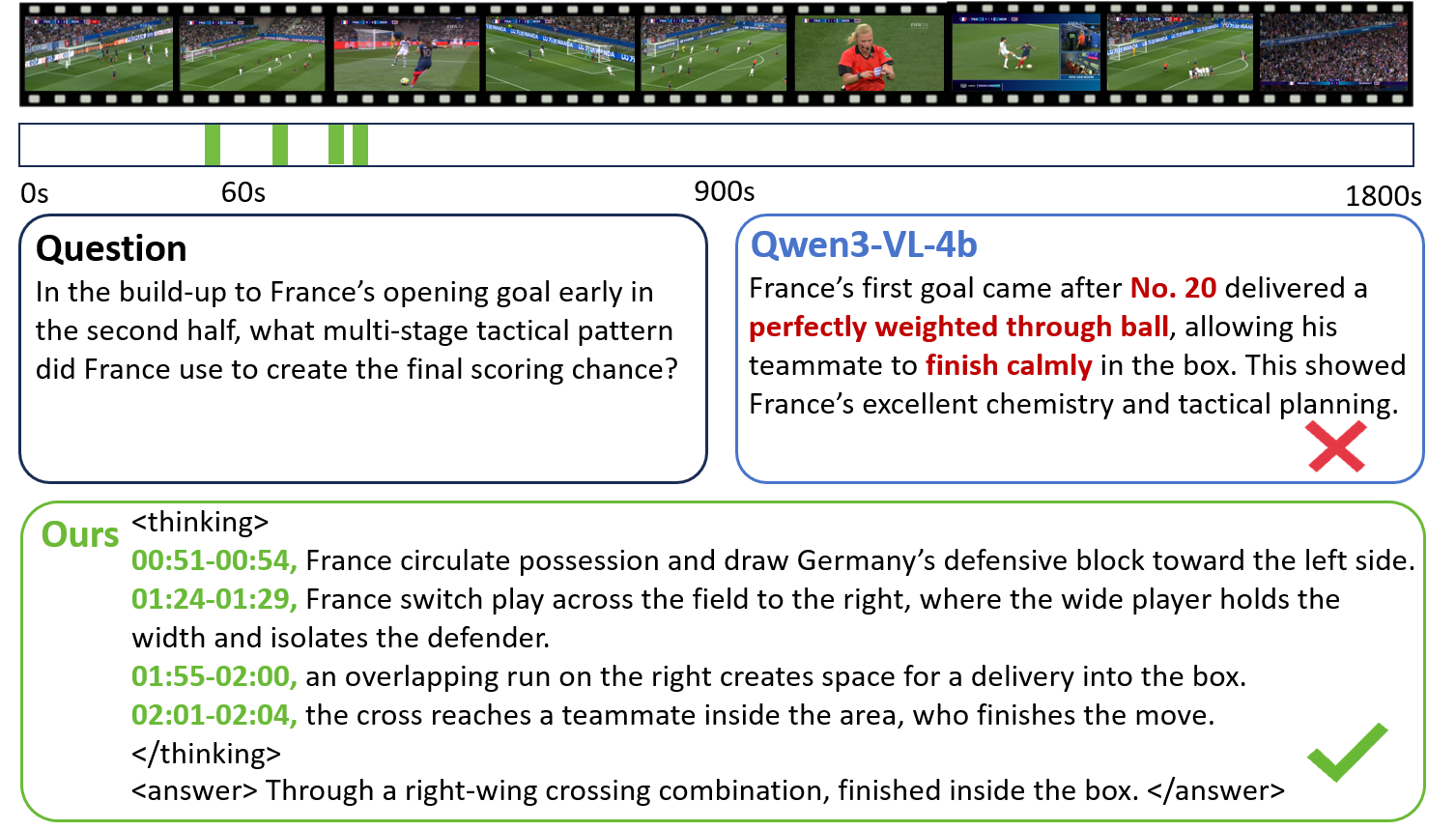}}
  \caption{\textbf{Chain-of-Time reasoning enables more reliable and verifiable answers in long-form sports videos.}}
  \label{fig:example}
\end{figure}

\section{Introduction}
\label{sec:intro}

Sports videos capture complex and dynamic human activities at scale, playing a central role in global cultural life and serving as a key data source for professional analytics. Over the past decade, artificial intelligence and computer vision techniques have fundamentally reshaped sports video analysis, enabling tasks such as player detection and tracking, action spotting, tactical analysis, etc~\cite{ghasemzadeh_deepsportlab_2021,cui_sportsmot_2023,deliege_soccernet-v2_2021,xuFineSportsMultipersonHierarchical2024}. More recently, the rapid development of Multimodal Large Language Models (MLLMs)~\cite{baiQwen3VLTechnicalReport2025a,clark_molmo2_2026,feng_video-r1_2025} has opened new opportunities toward unified sports video understanding, particularly for more diverse and open-ended tasks such as commentary generation~\cite{raoMatchTimeAutomaticSoccer2024a} and video question answering (VideoQA)~\cite{xiaSportQABenchmarkSports2024}, which require flexible and compositional inference over long-form multimodal content.

Despite significant advances, current MLLMs still struggle with long-horizon reasoning in videos~\cite{lu_elv-halluc_2025,zhangEventHallusionDiagnosingEvent2025,wangVideoHallucerEvaluatingIntrinsic2024,lee_noah_2025,rawal_argus_2025,li_vidhalluc_nodate}. This weakness is particularly evident in sports scenarios, which are long-form and highly dynamic, where interpreting sparse but critical events demands a holistic and long-horizon understanding. For example, in a soccer match, understanding how a goal is scored may require tracing a sequence of events, such as passes and player movements that unfold long before the final shot. 
Such scenarios expose a key limitation of current MLLMs: they struggle to identify temporally dispersed evidence and compose it into reasoning, a capability we refer to as temporal compositional reasoning.

We argue that this limitation mainly stems from 
two tightly coupled aspects:
(1) the scarcity of high-quality annotations that explicitly capture temporally dispersed evidence across multiple time spans, resulting in insufficient supervision for long-horizon reasoning~\cite{wang_lvbench_2025,chenCGbenchCluegroundedQuestion2024a}; and (2) the lack of methods that explicitly encourage models to identify, localize, and justify the specific temporal evidence underlying their final answers~\cite{liDiscoveringSpatiotemporalRationales2023}. As a result, models tend to overrely on language priors while underutilizing visual and temporal evidence, producing seemingly plausible yet weakly grounded answers and thus leading to hallucinations.

\begin{figure}[h]
  \centering
  \includegraphics[width=\linewidth]{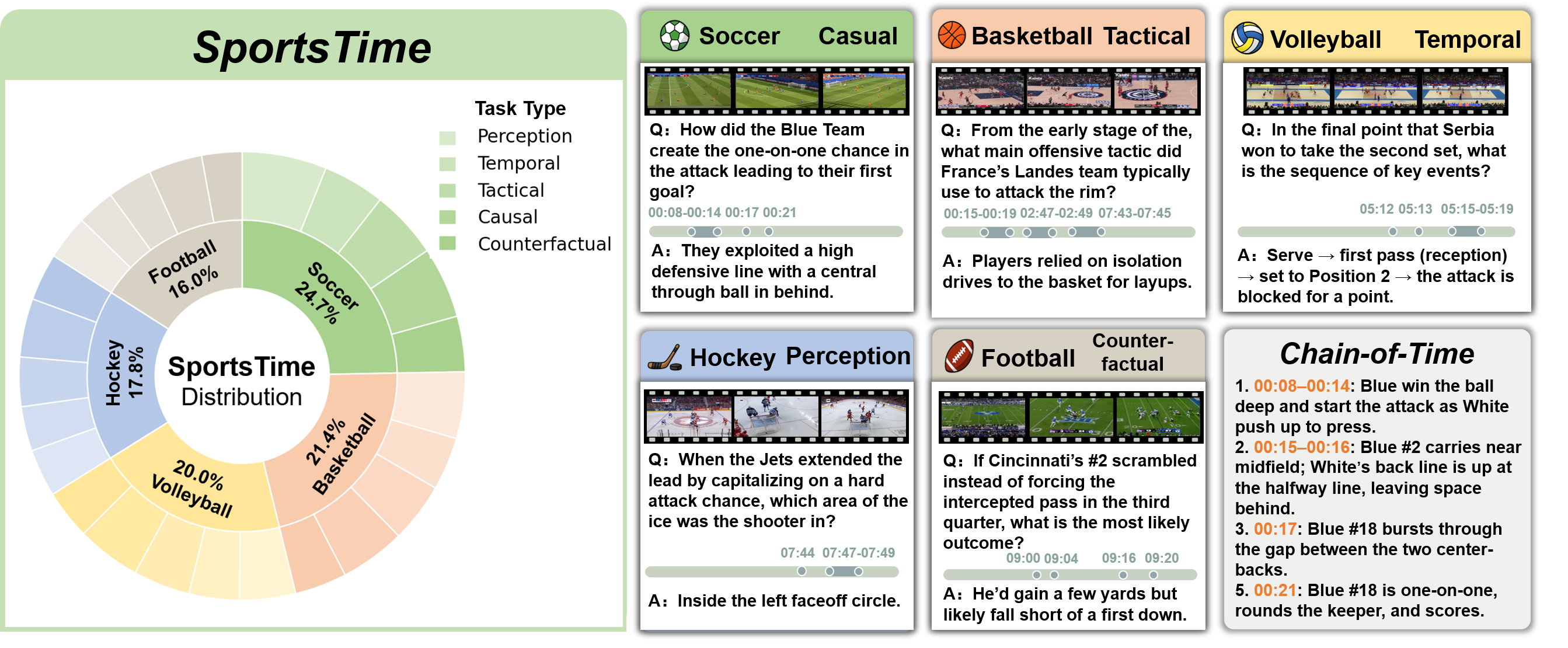}
  \caption{Overview of the SportsTime benchmark covering five sports and five reasoning types, with Chain-of-Time examples.}
  \label{fig:sportstime_overview}
\end{figure}

Motivated by this, we introduce \textbf{SportsTime}, a large-scale benchmark for comprehensive long-form sports video understanding through temporal compositional reasoning. SportsTime comprises over 14,000 high-quality open-ended QA pairs across five representative team sports, spanning both highlight videos ($\sim$ 10 min) and full-game broadcasts ($\sim$ 50 min). Distinct from existing datasets, most questions in SportsTime reflect real-world analytical demands, requiring models to integrate evidence from multiple temporally separated events rather than relying on a single moment. To support this, we further provide over 50,000 step-wise temporal evidence labels aligned with intermediate reasoning steps, enabling explicit supervision and fine-grained evaluation (see Fig.~\ref{fig:sportstime_overview} for data examples). We employ a dual-track evaluation protocol: an LLM-as-Judge framework for open-ended QA, and a specialized step-wise grounding alignment (SGA) evaluation to assess reasoning quality, with details provided in Section~\ref{sec:evaluation}.

To construct the benchmark at scale while maintaining domain fidelity, we adopt an expert-guided semi-automatic annotation pipeline. First, structured question templates for each sport are designed by domain experts, covering diverse tasks including perception, temporal, tactical, causal, and counterfactual reasoning. Then, these templates are instantiated by strong MLLMs to generate candidate QA samples and corresponding temporal evidence. Finally, peer reviewers and domain experts revise the annotations to ensure temporal accuracy, logical consistency, and domain correctness.

Building upon SportsTime, we further propose a \textbf{Chain-of-Time Reasoning (CoTR)} framework that formulates long-video understanding as a step-wise reasoning process over explicit temporal evidence. CoTR consists of two complementary mechanisms. First, during training, we introduce a temporal-reward Group Relative Policy Optimization (tr-GRPO) strategy that encourages models to align their reasoning steps with temporally grounded evidence, reinforcing correct localization and discouraging shortcuts. Second, during inference, we adopt a temporal search method based on an anchor-observe-infer loop. The model first anchors to candidate temporal segments, then iteratively observes retrieved clips to gather contextual evidence, and finally performs compositional reasoning to produce the final answer. This iterative evidence-seeking process enables explicit temporal grounding and improves long-horizon reasoning robustness.

Extensive experiments demonstrate that current MLLMs struggle with long-form compositional reasoning in the SportsTime benchmark, even strong proprietary models achieve limited accuracy. In contrast, the proposed CoTR method substantially improves both temporal localization and reasoning coherence by explicitly aligning intermediate reasoning steps with temporally grounded video evidence. These findings underscore the intrinsic difficulty of temporal compositional reasoning, demonstrating the necessity of benchmarks and methods that explicitly model dispersed temporal evidence to improve long-form video understanding. In summary,
our contributions are threefold:
\begin{itemize}[label=\textbullet]
    \item We introduce SportsTime, a large-scale benchmark for long-form sports video understanding, featuring temporal compositional reasoning tasks and fine-grained step-wise temporal evidence annotations.
    \item We propose Chain-of-Time Reasoning (CoTR), an evidence-driven method that enforces temporal grounding during training through reward-aligned optimization and enables robust long-horizon reasoning via an evidence-seeking inference loop.
    \item Extensive experiments demonstrate that SportsTime reveals substantial limitations of existing MLLMs in long-horizon reasoning, and that CoTR consistently improves both answer accuracy and step-wise temporal grounding alignment across diverse settings.
\end{itemize}

\section{Related Work}

\subsection{Multimodal Sports Benchmarks}
Existing multimodal sports benchmarks largely fall into two categories.
(1) Single-sport efforts, most notably in soccer, have progressed from task-specific benchmarks to more unified soccer-specific modeling. SoccerNet~\cite{deliege_soccernet-v2_2021} established representative tasks such as replay grounding and camera-shot segmentation, while later works including SoccerReplay-1988~\cite{raoUniversalSoccerVideo2025} and SoccerMaster~\cite{yangSoccerMasterVisionFoundation2025a} scale up multimodal soccer data and support more unified soccer understanding pipelines.
(2) Multi-sport benchmarks mainly broaden coverage across sports while moving from generic VideoQA toward richer perception and reasoning evaluation. Sports-QA~\cite{xiaSportQABenchmarkSports2024} and SPORTU~\cite{xiaSPORTUComprehensiveSports2025a} evaluate multi-sport QA, rule understanding, and slow-motion video reasoning, whereas SportR~\cite{xiaSportRBenchmarkMultimodal2025} introduces CoT and spatial-grounding supervision. FineSports~\cite{xuFineSportsMultipersonHierarchical2024} provides annotations for action understanding and spatio-temporal localization.
However, most benchmarks still emphasize short clips or fine-grained action labels, leaving long-form temporal compositional reasoning with step-wise verifiable evidence largely underexplored.

\subsection{Temporally Grounded Video Understanding}
Recent work has increasingly focused on temporal understanding in long videos.
One line of work~\cite{wuNumberItTemporal2025a,gupta_toga_2025,leonardis_timecraft_2025,wang_time-r1_2025,ding2026retrievingrelevantmomentsbenchmark} focuses on temporal grounding, aligning a language query with the relevant temporal segment in a video, i.e., query-to-time alignment, often by predicting timestamps, temporal spans, or moment proposals~\cite{chenCGbenchCluegroundedQuestion2024a,sugandhikaKnowshowBenchmarkingVideolanguage2025,guoVTGLLMIntegratingTimestamp,wangGroundedVideoLLMSharpeningFinegrained2025,yangTimeExpertExpertguidedVideo2025,wangVideoITGMultimodalVideo2025}. Another line of work~\cite{zhou_temporal_2018,li_temporal_2025,ye_re-thinking_2025,ren_testa_2023} investigates how temporally localized evidence can be used to support video understanding and reasoning.
TOGA~\cite{gupta_toga_2025} jointly predicts an open-ended answer and the temporal spans that support the final response. In addition, LongVT~\cite{yangLongVTIncentivizingThinking2025a} and VITAL~\cite{zhangThinkingVideosMultimodal2025} use temporal localization as part of a tool-augmented reasoning process over long videos. Unlike prior work, we formulate long-video understanding as chain-of-time reasoning, with step-wise temporal anchors serving as intermediate states that guide evidence acquisition and multi-step inference.

\section{SportsTime Benchmark}
\label{sec:benchmark}
This section introduces \textbf{SportsTime}, a long-form sports video understanding benchmark with fine-grained chain-of-time annotations, from its construction (Sec. \ref{sec:Dataset Construction}) and characteristics (Sec. \ref{sec:Dataset Characteristics}). Release details and ethical considerations are provided in Appendix~A.1.
\subsection{Benchmark Construction}
\label{sec:Dataset Construction}

\subsubsection{Benchmark Formulation and Design.} We formulate SportsTime as a long-form sports video reasoning benchmark with QA pairs and chain-of-time annotations. Each sample consists of a video, a question, an open-ended answer, and a step-wise reasoning chain in which each intermediate step is grounded in explicit temporal evidence. We highlight three key designs of SportsTime below.
(1) \textbf{Open-ended question answering.} We adopt open-ended QA rather than multiple-choice QA because it better reflects real-world usage and is also more challenging for models. (2) \textbf{Five task types for broad coverage.} We organize the questions into five task types, including causal, tactical, counterfactual, temporal, and perception reasoning. These categories cover a broad spectrum of abilities, from visual recognition to higher-level game understanding.
(3) \textbf{Fine-grained Chain-of-Time annotation.} A key feature of SportsTime is that we annotate each reasoning step with a supporting timestamp or time span, making the reasoning path explicitly verifiable and enabling direct supervision for temporally compositional long video reasoning.

\subsubsection{Data Collection.} To ensure diversity across sports, we collect official broadcast videos from five team sports, including American Football, Ice Hockey, Soccer, Basketball, and Volleyball, which differ substantially in camera conventions, editing styles, and event structures. In addition, we further include men's and women's games and international, professional, and collegiate events to introduce structured diversity in gameplay and presentation. SportsTime includes 1,575 videos, with 208 full-match videos and 1,367 highlight videos. We select full matches to benchmark long-horizon reasoning under sparse evidence, and highlights to improve coverage of event-dense and stylistically diverse scenarios.

\subsubsection{Annotation Pipeline.} 
\label{sec:Annotation Pipeline}
Since long-form sports video QA samples simultaneously require large-scale coverage, temporally valid evidence, and domain-correct sports knowledge, neither purely manual construction nor purely automatic generation is sufficient. 
To address this challenge, we propose \textbf{an expert-guided semi-automatic annotation framework}. After data collection, the framework consists of three stages: expert template design, candidate QA generation, and two-stage manual review.

\begin{figure}[h]
  \centering
  \includegraphics[width=\linewidth]{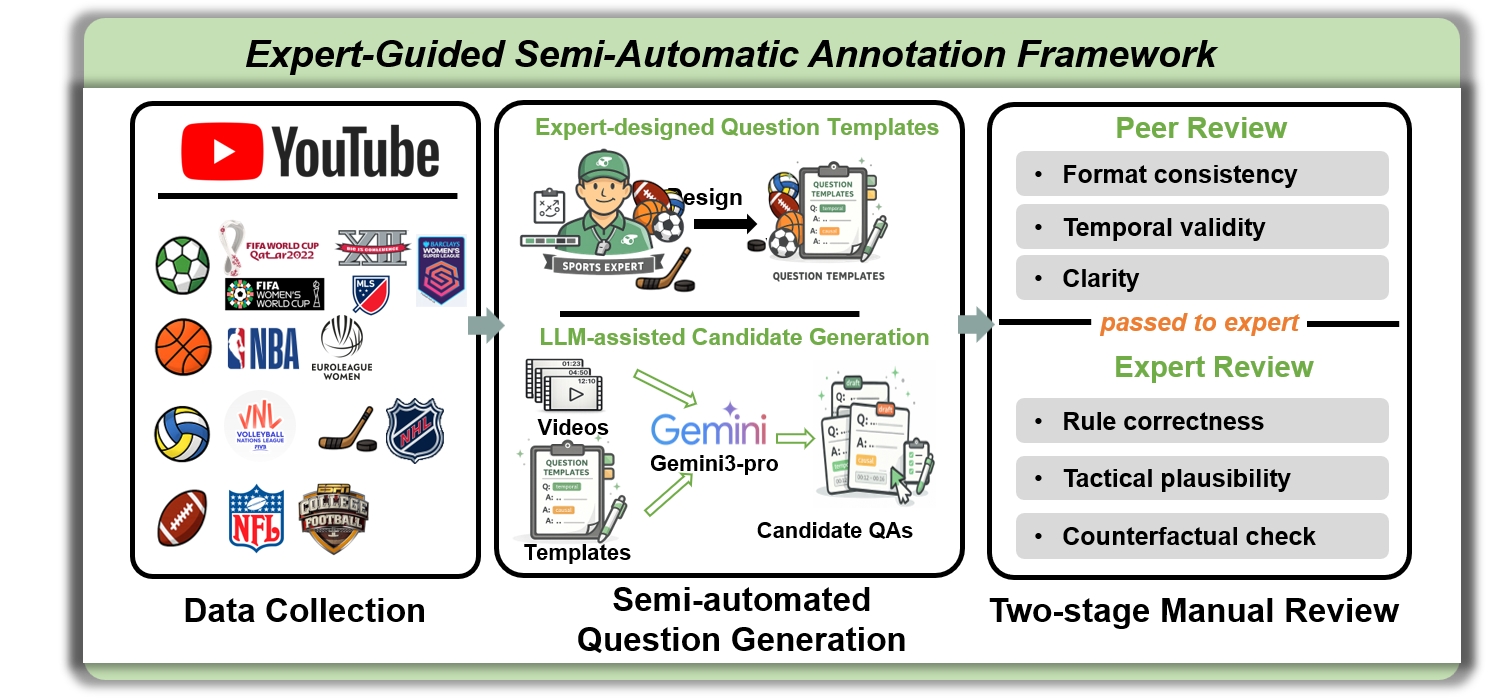}
  \caption{\textbf{Overview of our expert-guided semi-automatic annotation pipeline.} We collect long-form sports videos, generate candidate QAs with expert-designed templates and LLM assistance, and ensure quality through two-stage manual review.}
  \label{fig:placeholder}
\end{figure}

In the first stage, we invite sports experts to design question templates, which explicitly constrain the question types and semantic scope.
In the second stage, we use an LLM to generate candidate QA samples based on the videos and templates, which improves the efficiency of large-scale data construction. This design turns open-ended generation into controlled generation, improving the usability and coverage of the candidate samples.
Furthermore, to ensure data reliability, we introduce a two-stage manual review mechanism.
Firstly, peer annotators conduct an initial check, focusing on format consistency, temporal validity, and clarity of expression, in order to filter out obviously low-quality samples.
Secondly, domain experts perform final review, focusing on rule correctness, tactical plausibility, and the feasibility of counterfactual scenarios.
This pipeline preserves the scalability of sample generation while effectively reducing temporal errors, semantic ambiguity, and domain-knowledge bias.

\begin{table*}[h]
  \centering
  \begingroup
  \fontsize{6.8pt}{7.6pt}\selectfont
  \renewcommand{\arraystretch}{1.1}

  \newcommand{\STspan}{\textbf{S}}      % stepwise span
  \newcommand{\STtss}{\textbf{TS+S}}    % timestamp + short span
  \newcommand{\STnostep}{\textbf{--}}   % has time anno but not stepwise (if exists)
  \newcommand{\STna}{\xmark}            % no time annotation at all

  \caption{\textbf{Comparison with representative sports QA and video reasoning benchmarks.}
  \textbf{A} and \textbf{M} indicate automatic and manual annotation types, respectively.
  \textbf{Stepwise Time} indicates temporal grounding aligned to reasoning steps:
  \STspan = step-wise time spans, \STtss = step-wise timestamp and span.}
  \label{tab:dataset_comparison_compact}

  \resizebox{0.8\linewidth}{!}{%
  \begin{tabular}{@{}l c c r c c c c c@{}}
    \toprule
    \textbf{Dataset} &
    \textbf{Type} &
    \makecell[c]{\textbf{Dur.}\\\textbf{(s)}} &
    \makecell[c]{\textbf{\#QA}} &
    \makecell[c]{\textbf{Anno.}\\\textbf{Type}} &
    \textbf{CoT} &
    \makecell[c]{\textbf{Time}\\\textbf{Anno.}} &
    \makecell[c]{\textbf{Stepwise}\\\textbf{Time}} &
    \makecell[c]{\textbf{QA}\\\textbf{Type}} \\
    \midrule
    SportQA~\cite{xiaSportQABenchmarkSports2024}   & Sports    & clips   & 70,000 & A+M & \xmark & \xmark & \STna   & MCQ \\
    Sports-QA~\cite{liSportsQALargescaleVideo2024} & Sports    & 15.0    & 94,000 & A+M & \xmark & \xmark & \STna   & Open \\
    SPORTU~\cite{xiaSPORTUComprehensiveSports2025a} & Sports   & clips   & 12,048 & A+M & \cmark & \xmark & \STna   & MCQ+Open \\
    SPORTR~\cite{xiaSportRBenchmarkMultimodal2025} & Sports    & 4.96    & 20,000 & M   & \cmark  & \xmark & \STna   & MCQ+Open \\
    DeepSport~\cite{zouDeepSportMultimodalLarge2025} & Sports & -- & 84,700 & A & \cmark & \xmark & \STna & Open \\
    LongVideoBench~\cite{wuLongVideoBenchBenchmarkLongcontext2024} & General & 473.0  & 6,678 & M & \xmark & \xmark & \STna & MCQ \\
    LVBench~\cite{wang_lvbench_2025} & General & 4101.0 & 1,549 & M & \xmark & \cmark & \xmark & MCQ \\
    Video-MME\cite{fuVideoMMEFirstEverComprehensive2025a} & General & 1017.9 & 2,700 & M & \xmark & \xmark & \STna & MCQ \\
    MLVU\cite{zhouMLVUBenchmarkingMultitask2025}  & Narrative & 930     & 3,102 & M & \xmark & \xmark & \STna & MCQ \\
    MINERVA\cite{nagraniMINERVAEvaluatingComplex} & General   & 743.23  & 1,515 & M & \cmark & \cmark & \xmark & MCQ \\
    VRBench\cite{yuVRBenchBenchmarkMultiStep2025} & Narrative & 5,796.0 & 8,243 & M & \cmark & \cmark & \STspan & MCQ+Open \\
    \midrule
    \textbf{Ours} & Sports & 1053.25 & 14,326 & M & \cmark & \cmark & \STtss & Open \\
    \bottomrule
  \end{tabular}%
  }

  \endgroup
\end{table*}

\subsection{Benchmark Characteristics}
\label{sec:Dataset Characteristics}
\begin{figure}[h]
  \centering
  \includegraphics[width=\linewidth]{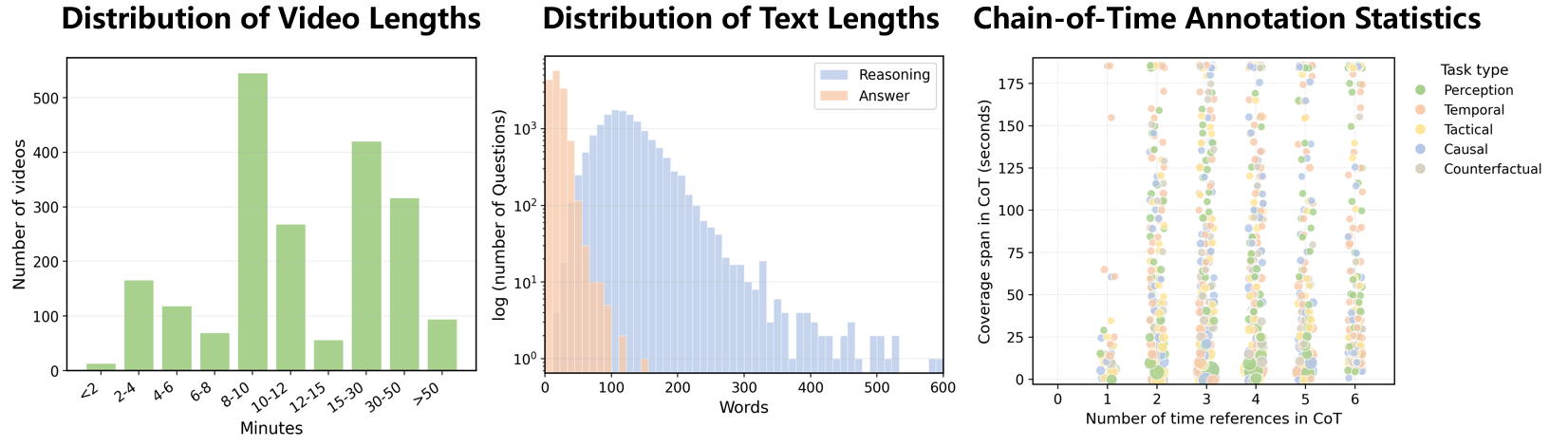}
  \caption{\textbf{Statistics of SportsTime.}
From left to right: video-length distribution, word-length distributions of reasoning chains and answers, and Chain-of-Time statistics.}
  \label{fig:distribution}
\end{figure}

Table \ref{tab:dataset_comparison_compact} compares SportsTime with sports QA and video understanding benchmarks, and Fig.~\ref{fig:distribution} summarizes the benchmark statistics of SportsTime.
Compared with existing sports QA datasets (e.g., SportQA~\cite{xiaSportQABenchmarkSports2024}), SportsTime targets longer videos and adopts an open-ended QA setting. This makes it better suited for evaluating compositional reasoning in realistic analysis scenarios.
Compared with prior benchmarks that include CoT annotations (e.g., SPORTU~\cite{xiaSPORTUComprehensiveSports2025a}, SPORTR~\cite{xiaSportRBenchmarkMultimodal2025}, and DeepSport~\cite{zouDeepSportMultimodalLarge2025}), SportsTime further provides step-wise temporal annotations. This design makes the reasoning process verifiable and enables direct supervision and evaluation of intermediate reasoning paths.
Compared with representative general long-video reasoning benchmarks (e.g., Video-MME, LongVideoBench, and MLVU, along with related long-video benchmarks~\cite{fuVideoMMEFirstEverComprehensive2025a,wuLongVideoBenchBenchmarkLongcontext2024,zhouMLVUBenchmarkingMultitask2025,wang_lvbench_2025,yuVRBenchBenchmarkMultiStep2025,nagraniMINERVAEvaluatingComplex}), SportsTime focuses on sports scenarios and more systematically captures domain-specific challenges, including long-horizon event evolution, sparse key events, and cross-segment evidence dependencies. In particular, compared with VRBench~\cite{yuVRBenchBenchmarkMultiStep2025}, our sports setting requires much finer temporal grounding at each reasoning step. Each step is typically associated with an event timestamp or a short, second-scale span, rather than the longer, minute-scale spans that often suffice in narrative videos. This reflects the sparsity and momentary nature of decisive sports events, which makes precise evidence composition substantially harder.
Overall, SportsTime provides a realistic and challenging testbed for broader research on long-video understanding.

\section{Method}
\begin{figure}[!t]
  \centering
  \includegraphics[width=0.9\linewidth]{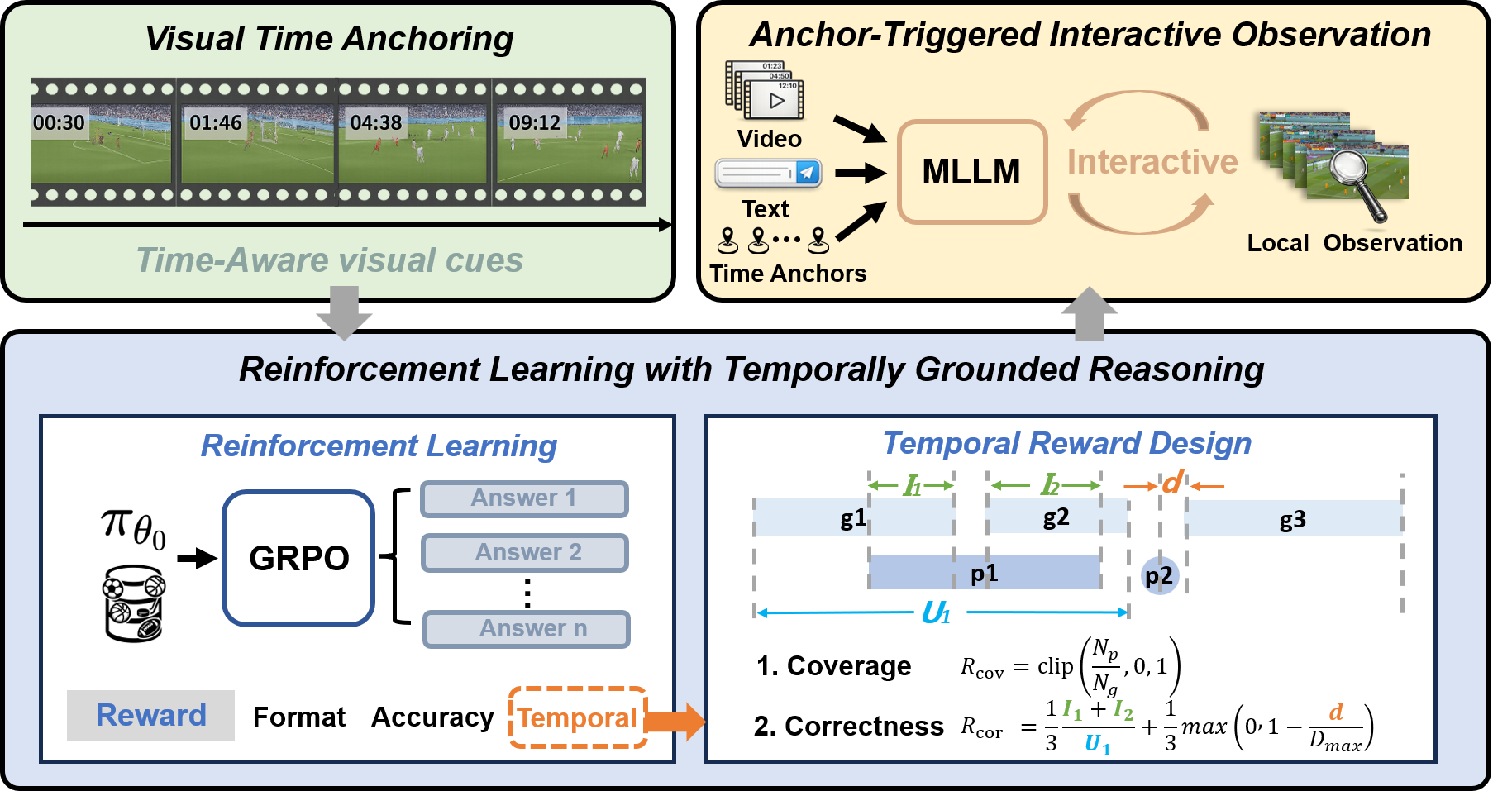}
  \caption{\textbf{Overview of the Chain-of-Time Reasoning (CoTR) Framework.} CoTR consists of three stages: visual time anchoring, reinforcement learning with temporally grounded reasoning, and anchor-triggered interactive observation.}
  \label{fig:method_overview}
\end{figure}

In this section, we present CoTR, our \textbf{C}hain-\textbf{o}f-\textbf{T}ime \textbf{R}easoning approach for temporal compositional reasoning. 
The core idea is to make the model reason with explicit temporal evidence. 
To reliably induce this behavior, we adopt a three-stage pipeline. 
First, we introduce a \textbf{visual time anchoring} step (Sec.~\ref{sec:method_timestamp}) that turns temporal grounding into a directly observable visual cue. 
Second, we perform \textbf{reinforcement learning with temporally grounded reasoning} (Sec.~\ref{sec:method_rl}) to encourage reasoning based on temporal evidence. 
Third, we introduce \textbf{anchor-triggered interactive observation} (Sec.~\ref{sec:method_inference}) to iteratively verify and revise reasoning via anchor-based local clip retrieval.
The overall framework of CoTR is illustrated in Fig.~\ref{fig:method_overview}.

\subsection{Problem Formulation}
CoTR frames long-form video QA as a chain-of-time reasoning problem. Given a video $V$ and a question $q$, the model predicts the answer $a$ through a sequence of reasoning steps, each grounded in localized temporal evidence. 
Concretely, we define each step as a tuple $n_t=\langle s_t,\tau_t\rangle$, where $s_t$ is a textual statement and $\tau_t$ is its supporting time anchor, represented as either a timestamp or a temporal span. 
The full trajectory becomes $\mathcal{T}=[n_1,\ldots,n_T,a]$, yielding the factorization
\begin{equation}
\pi_\theta(\mathcal{T}\mid V,q)=\Big(\prod_{t=1}^{T}\pi_\theta(n_t\mid V,q,n_{<t})\Big)\cdot \pi_\theta(a\mid V,q,n_{\le T}),
\end{equation}
Under this formulation, each intermediate claim is paired with a retrievable temporal anchor, which enables iterative local evidence verification and revision.

\subsection{Visual Time Anchoring}
\label{sec:method_timestamp}
A practical challenge is that many MLLMs struggle to align events with their exact times in long videos. Because temporal progress is not directly observable, temporal anchors can drift even when the event itself is correctly identified.
Inspired by the philosophy behind DeepSeekOCR~\cite{weiDeepSeekOCRContextsOptical2025} and Number it~\cite{wuNumberItTemporal2025a} that \emph{a picture is worth a thousand words}, we introduce a simple but effective method. We burn an explicit timestamp overlay into the top-right corner of video frames with a consistent \texttt{mm:ss} format.
This converts temporal grounding into a directly readable visual signal, allowing the model to infer time through visual-text recognition and reducing anchor drift in long contexts.
Empirically, this simple method improves the learnability of time anchoring.

\subsection{Reinforcement Learning with Temporally Grounded Reasoning}
\label{sec:method_rl}

To improve temporal grounding, we optimize the policy with reinforcement learning to promote evidence-seeking behavior and encourage the model to ground its reasoning on time anchors. Starting from the initial backbone policy $\pi_0$, we optimize a trajectory policy $\pi_\theta(\mathcal{T}\mid V,q)$ over anchored trajectories $\mathcal{T}=[n_1,\ldots,n_T,a]$ using temporal-reward GRPO objective, which improves stability by comparing rollouts within a group.
At each update, we sample multiple trajectories per $(V,q)$, compute rewards, form group-relative advantages, and update $\pi_\theta$ to maximize the reward while preserving the anchored generation protocol.

\subsubsection{Reward design.}
Our total reward is a weighted sum of three components,
\begin{equation}
R(\mathcal{T})=\lambda_{\text{fmt}}\,r_{\text{fmt}}(\mathcal{T})+\lambda_{\text{acc}}\,r_{\text{acc}}(\mathcal{T})+\lambda_{\text{temporal}}\,r_{\text{temporal}}(\mathcal{T}),
\end{equation}
where $r_{\text{fmt}}$ is a binary structural reward that checks whether the output follows the required format, and $r_{\text{acc}}$ is a task-level answer correctness reward. Our main focus here is $r_{\text{temporal}}$, which evaluates the quality of temporal grounding.
Concretely, our temporal reward consists of two parts—\emph{coverage} and \emph{correctness}:
\begin{equation}
r_{\text{temporal}}(\mathcal{T})=\alpha\,r_{\text{cov}}(\mathcal{T})+(1-\alpha)\,r_{\text{cor}}(\mathcal{T}),
\end{equation}
where $\alpha$ is a tunable weight.

\paragraph{(1) Step-wise coverage.}
$r_{\text{cov}}$ measures step-level temporal anchor coverage by rewarding trajectories in which reasoning steps are explicitly grounded in time.
Concretely, we segment the model's \texttt{<thinking>} into steps and compute the proportion of steps that contain at least one explicit time anchor.

\paragraph{(2) Temporal correctness.}
$r_{\text{cor}}$ measures how well predicted anchors align with ground-truth anchors.
We extract predicted anchors from the model's \texttt{<thinking>} and obtain ground-truth step anchors from the supervised process annotations.
For each ground-truth anchor, we compute its best-match score against the predicted anchors, using span IoU for span--span matches and a distance-aware similarity for point--span or point--point matches.
We then average these scores over all ground-truth anchors to obtain $r_{\text{cor}}\in[0,1]$.
Detailed matching rules and scoring definitions are provided in Appendix~B.2.

This reward design provides explicit supervision for temporal grounding:  $r_{\text{cov}}$ encourages the model to cover all required evidence, while $r_{\text{cor}}$ pushes predicted anchors toward the annotated evidence spans. As a result, tr-GRPO makes time anchors more learnable intermediate states, encouraging the policy to generate reasoning chains whose anchors are better aligned with verifiable evidence.

\subsection{Anchor-Triggered Interactive Observation}
\label{sec:method_inference}
Once the model has learned to produce time-anchored reasoning steps, we further use these anchors as actionable cues for test-time observation.
Given a video $V$ and a question $q$, the model generates an anchored reasoning trajectory
$\mathcal{T}=[n_1,\ldots,n_T,a]$ with $n_t=\langle s_t,\tau_t\rangle$,
where each step $s_t$ is accompanied by a temporal anchor $\tau_t$.
We treat each predicted $\tau_t$ as an explicit temporal query for local evidence retrieval in the corresponding neighborhood of $V$.

\subsubsection{Anchor-triggered local sampling.}
For a point anchor $\tau_t=\texttt{mm:ss}$, we sample a short temporal window centered at $\tau_t$; for a span anchor $\tau_t=[t_t^{s},t_t^{e}]$, we sample multiple clips uniformly from within the span.
Each anchor is thus converted into a set of local clips $\{c_{t}^{(j)}\}_{j=1}^{J_t}$, where each clip consists of $L$ frames sampled at a fixed stride.
This anchor-triggered sampling replaces global scanning of a long video with bounded local observation around predicted evidence locations.

\subsubsection{Evidence-grounded reasoning.}
We then perform reasoning over the anchored steps.
At turn $t$, the model is given the question $q$, the current step $s_t$, the anchor $\tau_t$, and the retrieved local clips $\{c_{t}^{(j)}\}_{j=1}^{J_t}$ as visual evidence.
Based on this local evidence, the model verifies and revises the current step before proceeding to the next turn.
Repeating this process yields a refined trajectory $\widetilde{\mathcal{T}}$ whose intermediate claims are explicitly checked against retrieved video content.
Finally, the model outputs the final answer $a$ conditioned on the accumulated verified evidence across turns.

\section{Experiments}
This section first introduces the experimental setting in Sec.~\ref{sec:Experimental setting}, and then presents the main results in Sec.~\ref{sec:Main Results}, followed by ablation studies in Sec.~\ref{sec:Ablation Studies}.
\subsection{Experimental Settings}
\label{sec:Experimental setting}
\subsubsection{Implementation Details.}
We sample 128 frames per video for training and up to 768 frames for inference. We use Qwen3-VL-4B as the backbone model. All experiments are conducted on 2× NVIDIA H100 (80GB). Additional implementation details, full GRPO hyperparameters, reward definitions, and the anchor-extraction parser used for Fig.~\ref{fig:cot_time_metrics}(\subref*{fig:cot_time_metrics_table})  are provided in Appendix~A.3.  
\subsubsection{Dual-Track Evaluation Protocol.}
\label{sec:evaluation}
We consider two settings for evaluation.
\paragraph{Open-ended QA.} We follow an LLM-as-Judge protocol and report all main results using a fixed judge (Qwen2.5-VL-7B~\cite{bai2025qwen25vltechnicalreport}). To assess the reliability of this, we further conduct a reliability study (Sec.~\ref{sec:Reliability of LLM-as-a-Judge.}) by comparing judgments from two additional LLM judges (MiniMax-M2.5 and GLM-4.7~\cite{teamGLM45AgenticReasoning2025}) and human raters.

\paragraph{Step-wise Grounding Alignment (SGA) Evaluation} We employ both objective and subjective assessments. Objectively, we measure temporal alignment to annotated evidence using temporal IoU between predicted time and ground-truth. Subjectively, we conduct human evaluation to assess (i) whether the evidence genuinely supports each reasoning step, (ii) whether intermediate conclusions are reasonable, and (iii) whether the overall chain is faithful and logically consistent.

\begin{table*}[t]
  \centering
  \renewcommand{\arraystretch}{1}
  \small
  \caption{\textbf{Performance on SportsTime.} ``Visual Input'' denotes the video input budget used for each model, reported either as the maximum number of sampled frames or the sampling rate. All scores are in \%.}
  \label{tab:SportsTime_results}

  \resizebox{0.95\linewidth}{!}{%
  \begin{tabular}{l c
                  S[table-format=3.2, round-mode=places, round-precision=2]
                  S[table-format=3.2, round-mode=places, round-precision=2]
                  S[table-format=3.2, round-mode=places, round-precision=2]
                  S[table-format=3.2, round-mode=places, round-precision=2]
                  S[table-format=3.2, round-mode=places, round-precision=2]
                  S[table-format=3.2, round-mode=places, round-precision=2]}
    \toprule
    \multirow{2}{*}{\textbf{Model}} &
    \multirow{2}{*}{\textbf{Visual Input}} &
    \multicolumn{6}{c}{\textbf{SportsTime}} \\
    \cmidrule(lr){3-8}
    & & \textbf{Perception} & \textbf{Temporal} & \textbf{Tactical} & \textbf{Causal} & \textbf{Counterfactual} & \textbf{Avg.} \\
    \midrule

    \rowcolor{groupbg}
    \multicolumn{8}{l}{\textbf{Proprietary}} \\
    GPT-5          & 0.2 fps & 38.77 & 27.45 & 43.85 & 44.77 & 46.93 & 40.72 \\
    Gemini-2.5-Pro & 0.2 fps & 39.66 & 19.78 & 29.49 & 39.41 & 13.71 & 29.37 \\
    \midrule

    \rowcolor{groupbg}
    \multicolumn{8}{l}{\textbf{Open-source}} \\
    Qwen3-VL-8B-Instruct~\cite{baiQwen3VLTechnicalReport2025a} & 768 frames& 24.27 & 14.31 & 33.26 & 23.08 & 44.47 & 27.45 \\
    Qwen3-VL-4B-Instruct & 768 frames& 23.26 & 13.14 & 31.11 & 22.92 & 34.17 & 25.15 \\
    VideoLLaMA3-7B~\cite{zhangVideoLLaMAInstructiontunedAudioVisual2023a} & 768 frames& 20.06 & 10.98 & 16.92 & 14.00 & 28.54 & 17.96 \\
    InternVideo2.5-8B~\cite{chen_expanding_2025} & 512 frames& 18.12 & 10.39 & 19.83 & 9.85 & 26.21 & 16.68 \\
    Video-R1-7B~\cite{feng_video-r1_2025} & 768 frames& 19.26 & 8.04 & 23.25 & 18.62 & 33.01 & 20.40 \\
    MiniCPM-V4.5-8B~\cite{yao2024minicpm} & 512 frames& 16.38 & 10.14 & 13.50 & 16.18 & 27.27 & 16.48 \\
    GLM-4.6v-Flash-9B~\cite{vteam2025glm45vglm41vthinkingversatilemultimodal} & 640 frames& 2.27 & 1.18 & 3.76 & 3.54 & 13.20 & 4.62 \\
    \bottomrule
  \end{tabular}%
  }
\end{table*}

\subsection{Main Results}
\label{sec:Main Results}
\subsubsection{Results on SportsTime.}
In Table~\ref{tab:SportsTime_results}, we report performance on SportsTime across five task types and the overall average. Overall, the absolute accuracy is still far from solved: even strong proprietary models reach only 40.72\% (GPT-5) and 29.37\% (Gemini-2.5-Pro), while open-source baselines are substantially lower. Such a low performance ceiling is expected because SportsTime demands temporal compositional reasoning over long-form matches, where decisive evidence is sparse, brief, and widely separated in time. Under a fixed frame budget, models may fail to retrieve the key moments and instead fall back to semantic priors, which particularly hurts Temporal and Causal questions that demand evidence-based linking across distant events.

\subsubsection{Effectiveness of CoTR.}
In Table~\ref{tab:cotr_results}, building on Qwen3-VL-4B, our CoTR yields consistent gains. \textbf{CoTR (full)} improves the overall average from 25.15\% to 29.24\%, with notable improvements on Counterfactual and Causal. 
After tr-GRPO fine-tuning, the model shows a pronounced gain on Tactical, which may primarily reflect improved command of sports-specific tactical concepts, and is potentially further aided by the temporal reward. With full CoTR, we improve both Perception (+3.16\%) and Causal (+4.29\%), suggesting that our method improves both the identification of relevant visual evidence and the integration of temporally dispersed clues for causal attribution. To validate the reliability of LLM-as-Judge, we additionally perform human evaluation on 200 stratified test samples, and find that its ranking is highly consistent with human judgments.

\begin{table*}[!t]
  \centering
  \renewcommand{\arraystretch}{1}
  \small
  \caption{\textbf{Effectiveness of CoTR on SportsTime.} All scores are in \%. Best and second-best results are highlighted by \Best{bold} and \Second{gray underline}, respectively. The last column reports \textbf{human evaluation} on a stratified subset of 200 test examples. \textbf{Improvement} denotes the absolute gain over \textit{the baseline model} (Qwen3-VL-4B-Instruct), measured in percentage points.}
  \label{tab:cotr_results}

  \resizebox{\textwidth}{!}{%
  \begin{tabular}{l
                  S[table-format=3.2, round-mode=places, round-precision=2]
                  S[table-format=3.2, round-mode=places, round-precision=2]
                  S[table-format=3.2, round-mode=places, round-precision=2]
                  S[table-format=3.2, round-mode=places, round-precision=2]
                  S[table-format=3.2, round-mode=places, round-precision=2]
                  S[table-format=3.2, round-mode=places, round-precision=2]
                  c}
    \toprule
    \multirow{2}{*}{\textbf{Model}} &
    \multicolumn{7}{c}{\textbf{SportsTime}} \\
    \cmidrule(lr){2-8}
    & \textbf{Perception} & \textbf{Temporal} & \textbf{Tactical} & \textbf{Causal} & \textbf{Counterfactual} & \textbf{Avg.} & \textbf{Human Avg.(subset)} \\
    \midrule
    Qwen3-VL-4B-Instruct   & \Second{23.26} & 13.14 & 31.11 & 22.92 & 34.17 & 25.15 & 24.50 \\
    Qwen2.5-VL-3B-Instruct & 16.67 & 7.06  & 27.18 & 12.31 & 22.52 & 17.16 & 17.50 \\
    InternVL2.5-4B~\cite{chen_expanding_2025} & 21.99 & 14.83 & 30.58 & 16.30 & 39.23 & 24.57 & 26.00 \\
    MiniCPM-v4-4B~\cite{yao2024minicpm}     & 14.21 & 10.36 & 19.20 & 12.08 & 26.06 & 16.23 & 16.50 \\
    Ovis2-4B~\cite{luOvis25TechnicalReport2025a} & 11.87 & 6.34  & 13.96 & 6.76  & 5.25  & 9.02  & 8.00 \\
    LLaVA-OneVision1.5-4B~\cite{anLLaVAOneVision15FullyOpen2025} & 13.88 & 7.40  & 13.61 & 7.73  & 16.57 & 11.81 & 12.50 \\
    \midrule
    \rowcolor{groupbg}
    \multicolumn{8}{l}{\textbf{Ours}} \\
    \textbf{Ours(tr-GRPO)} & 22.01 & \Second{13.73} & \Best{35.73} & \Second{25.54} & \Second{39.81} & \Second{27.31} & \Second{29.00} \\
    \textbf{Ours(full)}  & \Best{26.42} & \Best{15.43} & \Second{34.55} & \Best{27.21} & \Second{42.22} & \Best{29.24} & \Best{30.50} \\
    \rowcolor{blue!10}
    \textbf{Improvement}
    & \multicolumn{1}{c}{$(+3.16\%\uparrow)$}
    & \multicolumn{1}{c}{$(+2.29\%\uparrow)$}
    & \multicolumn{1}{c}{$(+3.44\%\uparrow)$}
    & \multicolumn{1}{c}{$(+4.29\%\uparrow)$}
    & \multicolumn{1}{c}{$(+8.05\%\uparrow)$}
    & \multicolumn{1}{c}{$(+4.09\%\uparrow)$}
    & \multicolumn{1}{c}{$(+6.00\%\uparrow)$} \\
    \bottomrule
  \end{tabular}%
  }
\end{table*}

\subsubsection{SGA Evaluation.}
\begin{figure*}[!t]
  \centering
  \renewcommand{\arraystretch}{1.10}

  % ---------------- Left: compact sub-table ----------------
\begin{subfigure}[h]{0.62\textwidth}
  \centering
  \caption{Accuracy and temporal evidence quality.}
  \label{fig:cot_time_metrics_table}
  \small
  \resizebox{\textwidth}{!}{%
  \begin{tabular}{l
                  S[table-format=2.2, round-mode=places, round-precision=2]
                  S[table-format=2.2, round-mode=places, round-precision=2]
                  S[table-format=1.4, round-mode=places, round-precision=4]
                  S[table-format=2.2, round-mode=places, round-precision=2]}
    \toprule
    \textbf{Model} &
    {\textbf{Acc}(\%)$\uparrow$} &
    {\textbf{Anchor}(\%)$\uparrow$} &
    {\textbf{mIoU}$\uparrow$} &
    {\textbf{H@0.5}(\%)$\uparrow$} \\
    \midrule

    \textit{Base} & 25.15 & 11.95 & 0.2601 & 30.03 \\
    Zero-shot CoT~\cite{kojimaLargeLanguageModels2023a} & 24.01 & 19.49 & 0.2042 & 26.34 \\
    Time-prompted CoT & 19.53 & 91.80 & 0.1244 & 12.12 \\
    \midrule
    Native-GRPO~\cite{shaoDeepSeekMathPushingLimits2024c}   & 26.72 & 13.10 & 0.3128 & 32.09 \\
    \textbf{tr-GRPO(Ours)} & 27.31 & 95.76 & 0.5812 & 56.89 \\
    \bottomrule
  \end{tabular}
  }
\end{subfigure}
  % ---------------- Right: sub-figure placeholder ----------------
  \begin{subfigure}[h]{0.37\textwidth}
  \centering
  \caption{Human assessment under SGA.}
  \label{fig:cot_time_metrics_plot}
  \includegraphics[width=1\linewidth]{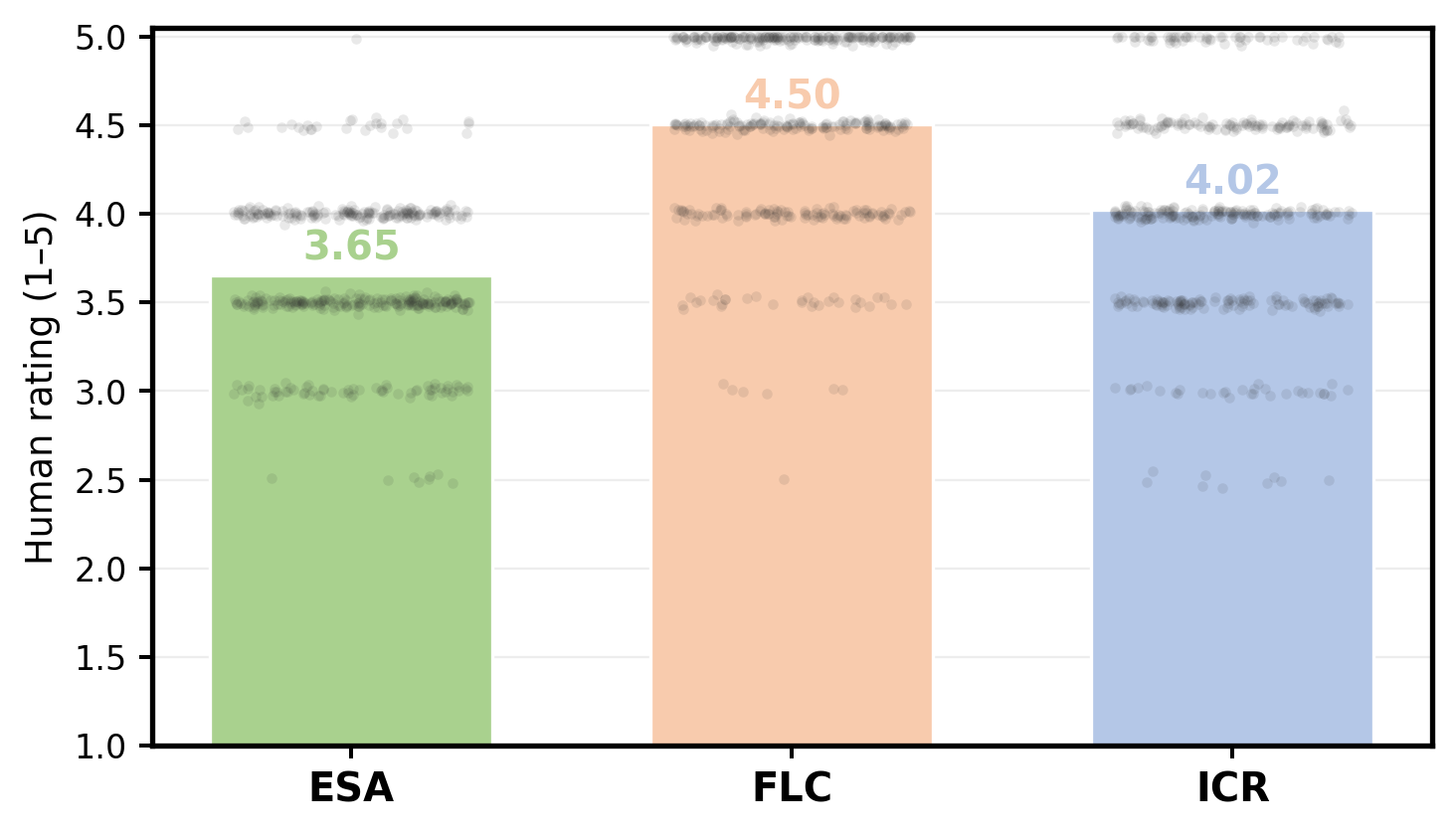}
  \end{subfigure}

  \caption{\textbf{SGA Evaluation.}    
      \footnotesize
       “mIoU” is the mean span IoU between predicted and reference time windows.
      “H@$\tau$” is the fraction of examples whose span IoU exceeds threshold $\tau$.
      Dots correspond to ratings of individual samples, while bars indicate the mean score.}
  \label{fig:cot_time_metrics}
\end{figure*}

We evaluate the quality of reasoning chains in two ways.
\paragraph{(1) Objective Evaluation.}
Fig.~\ref{fig:cot_time_metrics}(\subref*{fig:cot_time_metrics_table})  reports both final-answer accuracy and temporal evidence quality. Zero-shot CoT achieves only modest accuracy and, more importantly, rarely grounds its reasoning with explicit temporal spans (Anchor 19.49\%). In contrast, time-prompted CoT drastically increases the frequency of span outputs, but the spans are poorly aligned with the true evidence (mIoU 0.12, Hit@0.5 12.12\%). This gap indicates that simply “asking for timestamps” encourages format compliance rather than evidence faithfulness, i.e., the model may attach arbitrary or weakly related time windows to justify intermediate claims, resulting in temporal drift with superficially grounded rationales.

Our method closes this gap by explicitly internalizing the desired behavior into the reward design. In particular, tr-GRPO optimizes chain-of-time generation with rewards that jointly encourage temporal evidence coverage and correctness. Compared with native-GRPO, which improves answer accuracy but still exhibits weak temporal grounding, our reward design yields a fundamentally better trade-off between reasoning outcome and evidence quality. Ours reaches 27.31\% accuracy while maintaining high anchoring coverage (Anchor 95.76\%) and substantially stronger temporal alignment (mIoU 0.58, Hit@0.5 56.89\%). The large gains indicate that the improvement is not merely better answer matching, but a genuine reduction in temporal drift, as the predicted evidence more reliably support reasoning steps with the correct moments.

\paragraph{(2) Human Assessment under SGA.}
Fig.~\ref{fig:cot_time_metrics}(\subref*{fig:cot_time_metrics_plot}) reports human evaluation results. We randomly sample 100 examples and ask five raters to score the chains generated by CoTR on three dimensions: evidence-span accuracy (ESA), intermediate claim reliability (ICR), and faithful and logical consistency (FLC). CoTR achieves a high score on FLC (4.50/5), which indicates coherent and logically consistent chains, and a high ICR (4.02/5), suggesting largely reliable intermediate claims. ESA reaches 3.65/5, showing that although temporal spans generally support the corresponding steps, fine-grained localization remains improvable. 

\subsubsection{Comparison on Other Benchmarks.}
To evaluate generalization beyond SportsTime, we further test CoTR on several public long video reasoning benchmarks~\cite{wang_lvbench_2025,zhouMLVUBenchmarkingMultitask2025,fuVideoMMEFirstEverComprehensive2025a} in Table~\ref{tab:ext_benchmark_leaderboard}. CoTR achieves consistently competitive performance across these datasets, showing promising transferability beyond SportsTime.

\begin{table*}[!t]
  \centering
  % ---------- Left table ----------
  \begin{minipage}[h]{0.52\textwidth}
    \centering
    \renewcommand{\arraystretch}{0.6}
    \small
    \captionof{table}{\textbf{Comparison on general video benchmarks.} All scores are in \%. \textbf{Improvement} denotes the absolute gain over \textit{the baseline model} (Qwen3-VL-4B-Instruct).}
    \label{tab:ext_benchmark_leaderboard}
    \begin{adjustbox}{max width=\linewidth}
    \begin{tabular}{l
                    S[table-format=2.2]
                    S[table-format=2.2]
                    S[table-format=2.1]}
      \toprule
      \textbf{Models} & \textbf{LVBench} & \textbf{MLVU} & \textbf{VideoMME} \\
      \midrule
      Qwen3-VL-8B-Instruct & \Second{58.0} & \Best{78.1} & \Second{71.9} \\
      VideoLLaMA3-7B      & 44.3 & 68.7 & 61.1 \\
      InternVideo2.5-8B   & 46.4  & 72.8 & 65.1 \\
      Video-R1-7B         & 35.4 & 58.9 & 59.3 \\
      \midrule
      InternVL2.5-4B      & 40.75 & 48.3 & 63.6 \\
      SF-LLaVA-1.5-3B~\cite{xuSlowFastLLaVA15FamilyTokenefficient2025}      & 43.3 & 68.8 & 49.2 \\
      Qwen2.5-VL-3B-Instruct & 43.3 & 68.2 & 61.5 \\
      Qwen3-VL-4B-Instruct & 56.2 & 75.3 & 69.3 \\
      \rowcolor{groupbg}
      \textbf{Ours}  & \Best{59.9} & \Second{76.2} & \Best{72.1} \\
      \rowcolor{blue!12}
      \textbf{Improvement} 
      & \multicolumn{1}{c}{$(+3.7\%\uparrow)$}
      & \multicolumn{1}{c}{$(+0.9\%\uparrow)$}
      & \multicolumn{1}{c}{$(+2.8\%\uparrow)$}\\
      \bottomrule
    \end{tabular}
    \end{adjustbox}
  \end{minipage}
  \hspace{0.015\textwidth}
  % ---------- Right table ----------
  \begin{minipage}[h]{0.44\textwidth}
  \centering
  \renewcommand{\arraystretch}{0.88}
  \small
  \captionof{table}{\textbf{Ablation of our method.} All scores are in \%. The last two rows correspond to the main components of our method.}
  \label{tab:ablation_ours_subset}
  \begin{adjustbox}{max width=\linewidth}
  \begin{tabular}{
    l
    S[table-format=2.2, round-mode=places, round-precision=2]
    S[table-format=2.2, round-mode=places, round-precision=2]
    S[table-format=2.2, round-mode=places, round-precision=2]
  }
    \toprule
    \textbf{Setting} & \multicolumn{3}{c}{\textbf{SportsTime}} \\
    \cmidrule(lr){2-4}
    & \textbf{Temporal} & \textbf{Tactical} & \textbf{All} \\
    \midrule
    Base                & 13.14 & 31.11 & 25.15 \\
    + SFT               &  6.98 & 19.90 & 17.61 \\
    + Native-GRPO       & 11.76 & 32.65 & 26.72 \\
    + tr-GRPO w/o ts    & 13.14 & 32.21 & 26.12 \\
    \midrule
    + tr-GRPO w/ ts     & 13.73 & 35.73 & 27.31 \\
    \textbf{+ AT-IO (Ours)} & \textbf{15.43} & \textbf{34.55} & \textbf{29.24} \\
    \bottomrule
  \end{tabular}
  \end{adjustbox}
\end{minipage}
\end{table*}

\subsubsection{Reliability of LLM-as-Judge.}
\label{sec:Reliability of LLM-as-a-Judge.}
We use three independent LLM judges: GLM-4.7, MiniMax-M2.5, and Qwen2.5-VL-7B, together with human raters for open-ended evaluation on SportsTime.
As shown in Table~\ref{tab:judge_acc}, the judges achieve high average pairwise agreement, while Fleiss' $\kappa$ indicates moderate inter-judge consistency.
In addition, individual LLM judges show moderate agreement with human judgments.
These results support the use of LLM-as-Judge for scalable evaluation, although there remains room for improvement.

\subsection{Ablation Studies}
\label{sec:Ablation Studies}
\subsubsection{Method Ablation.}
We conduct ablations to assess our method (Table~\ref{tab:ablation_ours_subset}). A direct SFT baseline performs worse than the base model, with frequent repetitive generation on many samples. We attribute this to the structural variability and non-uniqueness of temporal reasoning chains, which make direct token-level imitation brittle and potentially harder to optimize at the 4B scale. Native-GRPO brings only a small gain and even reduces Temporal, suggesting that generic RL fine-tuning is insufficient. In contrast, tr-GRPO yields larger improvements, with a notable boost on Tactical, highlighting the role of our temporal-reward design. Interestingly, removing the timestamp-overlay (w/o ts) causes a clear drop, indicating that visually accessible timestamps help the model better leverage temporal evidence. Finally, adding AT-IO (\textbf{A}nchor-\textbf{T}riggered \textbf{I}nteractive \textbf{O}bservation) achieves the best overall performance, showing that test-time verification provides complementary benefits beyond reward-based training.

\begin{table*}[!t]
  \centering
  % ---------- Left: Judge table ----------
  \begin{minipage}[h]{0.42\textwidth}
    \centering
    \small
    \renewcommand{\arraystretch}{0.9}
    \captionof{table}{\textbf{Open-ended QA accuracy under multiple judges.} All scores are in \%.}
    \label{tab:judge_acc}
    \begin{adjustbox}{max width=\linewidth}
    \begin{tabular}{p{0.54\linewidth}cccc}
      \toprule
      \textbf{Model} &
      \shortstack{\textbf{Qwen}} &
      \textbf{MiniMax} &
      \textbf{GLM~\cite{teamGLM45AgenticReasoning2025}} &
      \textbf{Human} \\
      \midrule
      InternVideo2.5-8B & 16.68 & 14.25 & 17.94 & 17.37 \\
      Qwen3-VL-4B & 25.15 & 24.01 & 25.89 & 24.08 \\
      Ours-4B & 29.24 & 28.74 & 30.23 & 29.60 \\
      \midrule
    \multicolumn{5}{p{1.5\linewidth}}{\raggedright\footnotesize
    \textbf{\emph{Judge consistency:}} \\
    Avg.\ pairwise agreement = 88.34\%, Fleiss' $\kappa$ = 0.57.\\
    \textbf{\emph{Human alignment:}} \\
    Cohen's $\kappa$ (Qwen/MiniMax/GLM vs Human) = 0.6467/0.5759/0.5882.
    } \\
    \bottomrule

    \end{tabular}
    \end{adjustbox}
  \end{minipage}
  \hspace{0.02\textwidth}
  % ---------- Right: Two ablation figures ----------
  \begin{minipage}[h]{0.54\textwidth}
    \centering

    \begin{subfigure}[h]{0.49\linewidth}
      \centering
      \includegraphics[width=\linewidth]{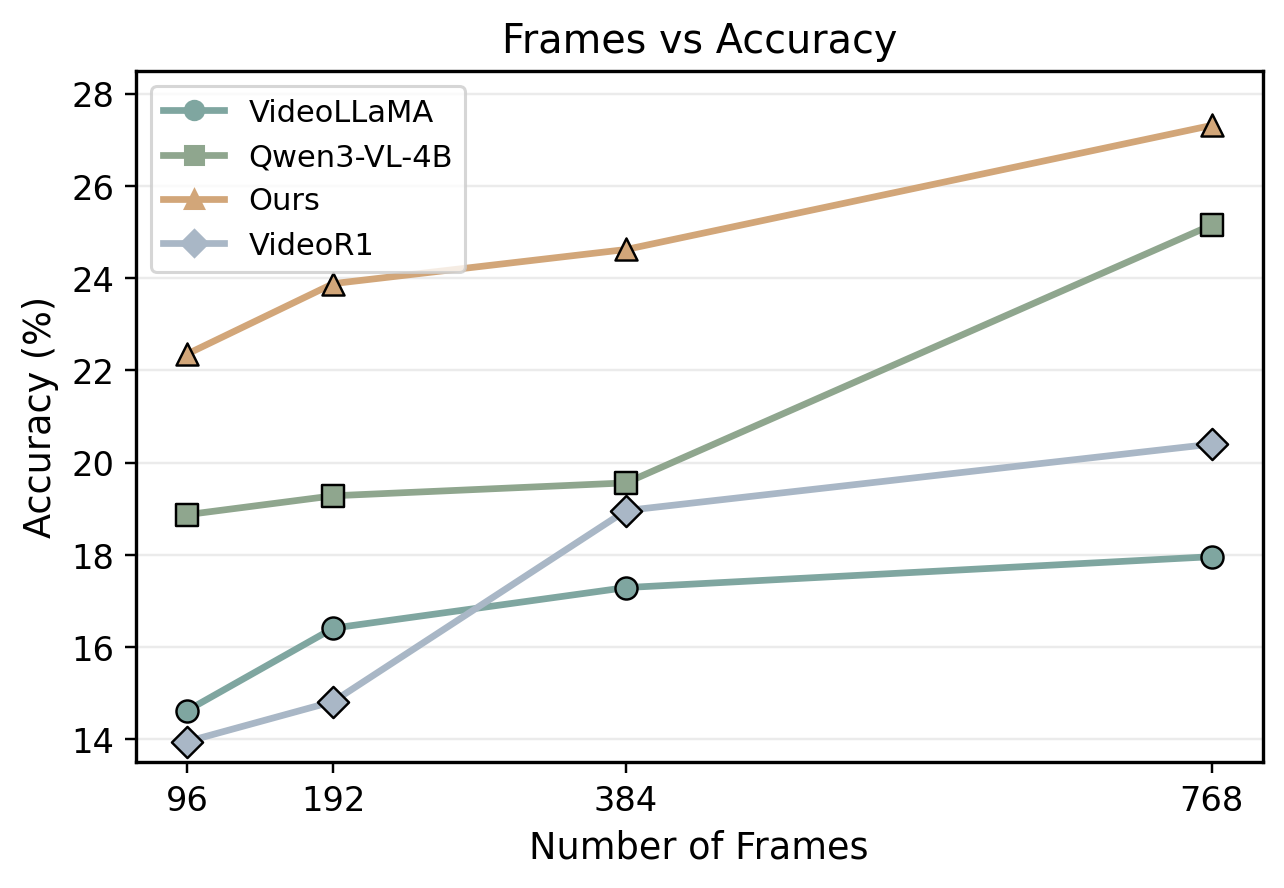}
      \caption{Frames vs. Acc.}
      \label{fig:video_frames}
    \end{subfigure}
    \hfill
    \begin{subfigure}[h]{0.49\linewidth}
      \centering
      \includegraphics[width=\linewidth]{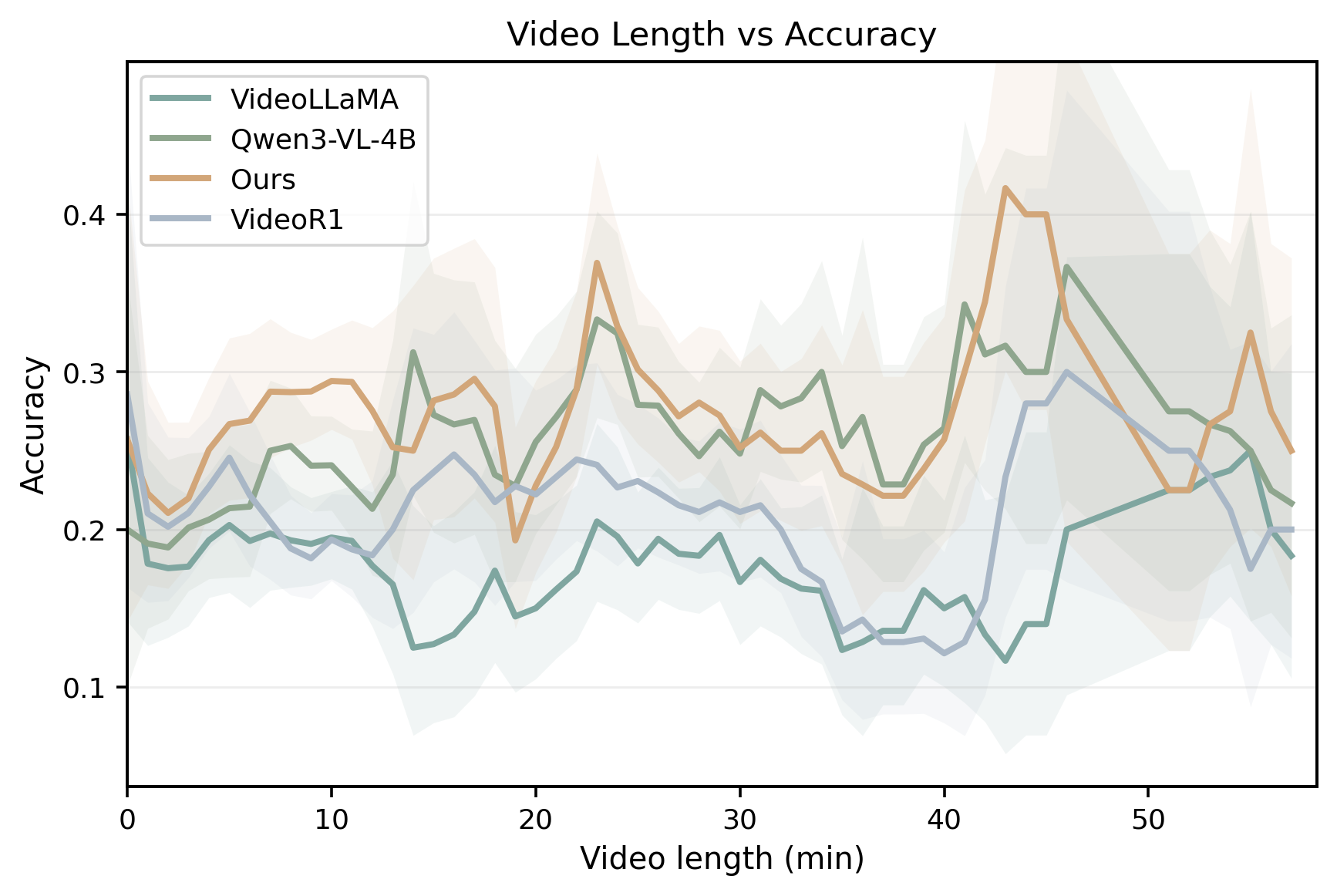}
      \caption{Video length vs. Acc.}
      \label{fig:video_length}
    \end{subfigure}
    \captionof{figure}{\textbf{Video setting ablation studies.}
    (a) Accuracy as a function of frame budget. (b) Accuracy as a function of video length.}
    \label{fig:ablations_row}
  \end{minipage}
\end{table*}
\subsubsection{Video Setting Ablation.}
Fig.~\ref{fig:ablations_row}(\subref*{fig:video_frames}) shows that accuracy generally improves with larger frame budgets, but the gain is model-dependent.
Fig.~\ref{fig:ablations_row}(\subref*{fig:video_length}) reveals a surprising pattern: accuracy does not decrease monotonically with video length, so length alone is not the decisive factor. Instead, difficulty is likely driven by confounding factors such as video type and event structure. Notably, our method retains an advantage in longer videos, indicating its effectiveness.

\section{Conclusion}
In this work, we introduce \textbf{SportsTime}, a large-scale benchmark for long-form sports video understanding with step-wise temporal evidence annotations, and propose \textbf{Chain-of-Time Reasoning (CoTR)}, a framework that unifies temporally grounded training and inference for long video reasoning. Our results show that explicit temporal evidence supervision and step-wise evidence-seeking substantially improve both temporal compositional reasoning and grounding quality. 
We hope SportsTime will provide a strong testbed for future research on long-horizon, evidence-grounded video understanding.

\section*{Acknowledgements}
This research is supported by Key Science \& Technology Project of Anhui Province under Grant No. 202523j08050027 and the National Key Research and Development Program of China under Grant No. 2025YFE0216600.

\bibliographystyle{splncs04}
\bibliography{main}

@String(CVPR  = {IEEE Conf. Comput. Vis. Pattern Recog.})

@String(ICCV  = {Int. Conf. Comput. Vis.})

@String(ECCV  = {Eur. Conf. Comput. Vis.})

@String(NeurIPS = {Adv. Neural Inform. Process. Syst.})

@String(CVPRW = {IEEE Conf. Comput. Vis. Pattern Recog. Worksh.})

@String(AAAI  = {AAAI})

@String(CVPR  = {CVPR})

@String(ICCV  = {ICCV})

@String(ECCV  = {ECCV})

@String(NeurIPS = {NeurIPS})

@String(CVPRW = {CVPRW})

@article{baiQwen3VLTechnicalReport2025a,
  title={Qwen3-vl technical report},
  author={Bai, Shuai and Cai, Yuxuan and Chen, Ruizhe and Chen, Keqin and Chen, Xionghui and Cheng, Zesen and Deng, Lianghao and Ding, Wei and Gao, Chang and Ge, Chunjiang and others},
  journal={arXiv preprint arXiv:2511.21631},
  year={2025}
}

@article{chenCGbenchCluegroundedQuestion2024a,
  title={Cg-bench: Clue-grounded question answering benchmark for long video understanding},
  author={Chen, Guo and Liu, Yicheng and Huang, Yifei and He, Yuping and Pei, Baoqi and Xu, Jilan and Wang, Yali and Lu, Tong and Wang, Limin},
  journal={arXiv preprint arXiv:2412.12075},
  year={2024}
}

@article{chen_expanding_2025,
  title={Expanding performance boundaries of open-source multimodal models with model, data, and test-time scaling},
  author={Chen, Zhe and Wang, Weiyun and Cao, Yue and Liu, Yangzhou and Gao, Zhangwei and Cui, Erfei and Zhu, Jinguo and Ye, Shenglong and Tian, Hao and Liu, Zhaoyang and others},
  journal={arXiv preprint arXiv:2412.05271},
  year={2024}
}

@InProceedings{fuVideoMMEFirstEverComprehensive2025a,
    author    = {Fu, Chaoyou and Dai, Yuhan and Luo, Yongdong and Li, Lei and Ren, Shuhuai and Zhang, Renrui and Wang, Zihan and Zhou, Chenyu and Shen, Yunhang and Zhang, Mengdan and Chen, Peixian and Li, Yanwei and Lin, Shaohui and Zhao, Sirui and Li, Ke and Xu, Tong and Zheng, Xiawu and Chen, Enhong and Shan, Caifeng and He, Ran and Sun, Xing},
    title     = {Video-MME: The First-Ever Comprehensive Evaluation Benchmark of Multi-modal LLMs in Video Analysis},
    booktitle = CVPR,
    month     = {June},
    year      = {2025},
    pages     = {24108-24118}
}

@article{guoVTGLLMIntegratingTimestamp, 
title={VTG-LLM: Integrating Timestamp Knowledge into Video LLMs for Enhanced Video Temporal Grounding}, 
volume={39}, 
number={3}, 
journal=AAAI, 
author={Guo, Yongxin and Liu, Jingyu and Li, Mingda and Cheng, Dingxin and Tang, Xiaoying and Sui, Dianbo and Liu, Qingbin and Chen, Xi and Zhao, Kevin}, 
year={2025}, 
month={Apr.}, 
pages={3302-3310} }

@article{liSportsQALargescaleVideo2024,
  title={Sports-qa: A large-scale video question answering benchmark for complex and professional sports},
  author={Li, Haopeng and Deng, Andong and Ke, Qiuhong and Liu, Jun and Rahmani, Hossein and Guo, Yulan and Schiele, Bernt and Chen, Chen},
  journal={arXiv preprint arXiv:2401.01505},
  year={2024}
}

@InProceedings{nagraniMINERVAEvaluatingComplex,
    author    = {Nagrani, Arsha and Menon, Sachit and Iscen, Ahmet and Buch, Shyamal and Mehran, Ramin and Jha, Nilpa and Hauth, Anja and Zhu, Yukun and Vondrick, Carl and Sirotenko, Mikhail and Schmid, Cordelia and Weyand, Tobias},
    title     = {MINERVA: Evaluating Complex Video Reasoning},
    booktitle = ICCV,
    month     = {October},
    year      = {2025},
    pages     = {23968-23978}
}

@InProceedings{raoUniversalSoccerVideo2025,
    author    = {Rao, Jiayuan and Wu, Haoning and Jiang, Hao and Zhang, Ya and Wang, Yanfeng and Xie, Weidi},
    title     = {Towards Universal Soccer Video Understanding},
    booktitle = CVPR,
    month     = {June},
    year      = {2025},
    pages     = {8384-8394}
}

@article{sugandhikaKnowshowBenchmarkingVideolanguage2025,
  title={Know-Show: Benchmarking Video-Language Models on Spatio-Temporal Grounded Reasoning},
  author={Sugandhika, Chinthani and Li, Chen and Rajan, Deepu and Fernando, Basura},
  journal={arXiv preprint arXiv:2512.05513},
  year={2025}
}

@article{wangGroundedVideoLLMSharpeningFinegrained2025,
  title={Grounded-videollm: Sharpening fine-grained temporal grounding in video large language models},
  author={Wang, Haibo and Xu, Zhiyang and Cheng, Yu and Diao, Shizhe and Zhou, Yufan and Cao, Yixin and Wang, Qifan and Ge, Weifeng and Huang, Lifu},
  journal={arXiv preprint arXiv:2410.03290},
  year={2024}
}

@article{wangVideoITGMultimodalVideo2025,
  title={VideoITG: Multimodal Video Understanding with Instructed Temporal Grounding},
  author={Wang, Shihao and Chen, Guo and Huang, De-an and Li, Zhiqi and Li, Minghan and Li, Guilin and Alvarez, Jose M and Zhang, Lei and Yu, Zhiding},
  journal={arXiv preprint arXiv:2507.13353},
  year={2025}
}

@article{weiDeepSeekOCRContextsOptical2025,
  title={Deepseek-ocr: Contexts optical compression},
  author={Wei, Haoran and Sun, Yaofeng and Li, Yukun},
  journal={arXiv preprint arXiv:2510.18234},
  year={2025}
}

@InProceedings{wuLongVideoBenchBenchmarkLongcontext2024,
  title={Longvideobench: A benchmark for long-context interleaved video-language understanding},
  author={Wu, Haoning and Li, Dongxu and Chen, Bei and Li, Junnan},
  booktitle=NeurIPS,
  volume={37},
  pages={28828--28857},
  year={2024}
}

@InProceedings{wuNumberItTemporal2025a,
    author    = {Wu, Yongliang and Hu, Xinting and Sun, Yuyang and Zhou, Yizhou and Zhu, Wenbo and others},
    title     = {Number it: Temporal Grounding Videos like Flipping Manga},
    booktitle = CVPR,
    month     = {June},
    year      = {2025},
    pages     = {13754-13765}
}

@inproceedings{xiaSportQABenchmarkSports2024,
  title={Sportqa: A benchmark for sports understanding in large language models},
  author={Xia, Haotian and Yang, Zhengbang and Wang, Yuqing and Tracy, Rhys and Zhao, Yun and Huang, Dongdong and Chen, Zezhi and Zhu, Yan and Wang, Yuan-fang and Shen, Weining},
  booktitle={Proceedings of the 2024 Conference of the North American Chapter of the Association for Computational Linguistics: Human Language Technologies (Volume 1: Long Papers)},
  pages={5061--5081},
  year={2024}
}

@article{xiaSportRBenchmarkMultimodal2025,
  title={SportR: A Benchmark for Multimodal Large Language Model Reasoning in Sports},
  author={Xia, Haotian and Ge, Haonan and Zou, Junbo and Choi, Hyun Woo and Zhang, Xuebin and Suradja, Danny and Rui, Botao and Tran, Ethan and Jin, Wendy and Ye, Zhen and others},
  journal={arXiv preprint arXiv:2511.06499},
  year={2025}
}

@article{xiaSPORTUComprehensiveSports2025a,
  title={Sportu: A comprehensive sports understanding benchmark for multimodal large language models},
  author={Xia, Haotian and Yang, Zhengbang and Zou, Junbo and Tracy, Rhys and Wang, Yuqing and Lu, Chi and Lai, Christopher and He, Yanjun and Shao, Xun and Xie, Zhuoqing and others},
  journal={arXiv preprint arXiv:2410.08474},
  year={2024}
}

@inproceedings{xuFineSportsMultipersonHierarchical2024,
  title = {{{FineSports}}: {{A}} Multi-Person Hierarchical Sports Video Dataset for Fine-Grained Action Understanding},
  shorttitle = {{{FineSports}}},
  booktitle = CVPR,
  author = {Xu, Jinglin and Zhao, Guohao and Yin, Sibo and Zhou, Wenhao and Peng, Yuxin},
  year = 2024,
  month = jun,
  pages = {21773--21782},
  publisher = {IEEE},
  address = {Seattle, WA, USA},
  urldate = {2025-11-08},
  isbn = {979-8-3503-5300-6}
}

@article{yangLongVTIncentivizingThinking2025a,
  title={Longvt: Incentivizing" thinking with long videos" via native tool calling},
  author={Yang, Zuhao and Wang, Sudong and Zhang, Kaichen and Wu, Keming and Leng, Sicong and Zhang, Yifan and Li, Bo and Qin, Chengwei and Lu, Shijian and Li, Xingxuan and others},
  journal={arXiv preprint arXiv:2511.20785},
  year={2025}
}

@article{yangSoccerMasterVisionFoundation2025a,
  title={SoccerMaster: A Vision Foundation Model for Soccer Understanding},
  author={Yang, Haolin and Rao, Jiayuan and Wu, Haoning and Xie, Weidi},
  journal={arXiv preprint arXiv:2512.11016},
  year={2025}
}

@InProceedings{yangTimeExpertExpertguidedVideo2025,
    author    = {Yang, Zuhao and Yu, Yingchen and Zhao, Yunqing and Lu, Shijian and Bai, Song},
    title     = {TimeExpert: An Expert-Guided Video LLM for Video Temporal Grounding},
    booktitle = ICCV,
    month     = {October},
    year      = {2025},
    pages     = {24286-24296}
}

@InProceedings{yuVRBenchBenchmarkMultiStep2025,
    author    = {Yu, Jiashuo and Wu, Yue and Chu, Meng and Ren, Zhifei and Huang, Zizheng and Chu, Pei and Zhang, Ruijie and He, Yinan and Li, Qirui and Li, Songze and Li, Zhenxiang and Tu, Zhongying and He, Conghui and Qiao, Yu and Wang, Yali and Wang, Yi and Wang, Limin},
    title     = {VRBench: A Benchmark for Multi-Step Reasoning in Long Narrative Videos},
    booktitle = ICCV,
    month     = {October},
    year      = {2025},
    pages     = {21655-21666}
}

@article{zhangThinkingVideosMultimodal2025,
  title={Thinking with videos: Multimodal tool-augmented reinforcement learning for long video reasoning},
  author={Zhang, Haoji and Gu, Xin and Li, Jiawen and Ma, Chixiang and Bai, Sule and Zhang, Chubin and Zhang, Bowen and Zhou, Zhichao and He, Dongliang and Tang, Yansong},
  journal={arXiv preprint arXiv:2508.04416},
  year={2025}
}

@InProceedings{zhouMLVUBenchmarkingMultitask2025,
    author    = {Zhou, Junjie and Shu, Yan and Zhao, Bo and Wu, Boya and Liang, Zhengyang and others},
    title     = {MLVU: Benchmarking Multi-task Long Video Understanding},
    booktitle = CVPR,
    month     = {June},
    year      = {2025},
    pages     = {13691-13701}
}

@article{zouDeepSportMultimodalLarge2025,
  title={DeepSport: A Multimodal Large Language Model for Comprehensive Sports Video Reasoning via Agentic Reinforcement Learning},
  author={Zou, Junbo and Xia, Haotian and Ye, Zhen and Zhang, Shengjie and Lai, Christopher and Ordonez, Vicente and Shen, Weining and Chen, Hanjie},
  journal={arXiv preprint arXiv:2511.12908},
  year={2025}
}

@inproceedings{leonardis_timecraft_2025,
  title={Timecraft: Navigate weakly-supervised temporal grounded video question answering via bi-directional reasoning},
  author={Liu, Huabin and Ma, Xiao and Zhong, Cheng and Zhang, Yang and Lin, Weiyao},
  booktitle=ECCV,
  pages={92--107},
  year={2024},
  organization={Springer}
}

@InProceedings{gupta_toga_2025,
    author    = {Gupta, Ayush and Roy, Anirban and Chellappa, Rama and Bastian, Nathaniel D. and Velasquez, Alvaro and Jha, Susmit},
    title     = {TOGA: Temporally Grounded Open-Ended Video QA with Weak Supervision},
    booktitle = ICCV,
    month     = {October},
    year      = {2025},
    pages     = {23593-23603}
}

@article{feng_video-r1_2025,
  title={Video-r1: Reinforcing video reasoning in mllms},
  author={Feng, Kaituo and Gong, Kaixiong and Li, Bohao and Guo, Zonghao and Wang, Yibing and Peng, Tianshuo and Wu, Junfei and Zhang, Xiaoying and Wang, Benyou and Yue, Xiangyu},
  journal={arXiv preprint arXiv:2503.21776},
  year={2025}
}

@article{clark_molmo2_2026,
  title={Molmo2: Open Weights and Data for Vision-Language Models with Video Understanding and Grounding},
  author={Clark, Christopher and Zhang, Jieyu and Ma, Zixian and Park, Jae Sung and Salehi, Mohammadreza and Tripathi, Rohun and Lee, Sangho and Ren, Zhongzheng and Kim, Chris Dongjoo and Yang, Yinuo and others},
  journal={arXiv preprint arXiv:2601.10611},
  year={2026}
}

@inproceedings{zhangVideoLLaMAInstructiontunedAudioVisual2023a,
    title = "Video-{LL}a{MA}: An Instruction-tuned Audio-Visual Language Model for Video Understanding",
    author = "Zhang, Hang  and
      Li, Xin  and
      Bing, Lidong",
    booktitle = "Proceedings of the 2023 Conference on Empirical Methods in Natural Language Processing: System Demonstrations",
    month = dec,
    year = "2023",
    pages = "543--553",
}

@article{wangVideoHallucerEvaluatingIntrinsic2024,
  title={Videohallucer: Evaluating intrinsic and extrinsic hallucinations in large video-language models},
  author={Wang, Yuxuan and Wang, Yueqian and Zhao, Dongyan and Xie, Cihang and Zheng, Zilong},
  journal={arXiv preprint arXiv:2406.16338},
  year={2024}
}

@article{zhangEventHallusionDiagnosingEvent2025,
  title={Eventhallusion: Diagnosing event hallucinations in video llms},
  author={Zhang, Jiacheng and Jiao, Yang and Chen, Shaoxiang and Zhao, Na and Tan, Zhiyu and others},
  journal={arXiv preprint arXiv:2409.16597},
  year={2024}
}

@article{ghasemzadeh_deepsportlab_2021,
  title={Deepsportlab: a unified framework for ball detection, player instance segmentation and pose estimation in team sports scenes},
  author={Ghasemzadeh, Seyed Abolfazl and Van Zandycke, Gabriel and Istasse, Maxime and Sayez, Niels and Moshtaghpour, Amirafshar and De Vleeschouwer, Christophe},
  journal={arXiv preprint arXiv:2112.00627},
  year={2021}
}

@InProceedings{cui_sportsmot_2023,
    author    = {Cui, Yutao and Zeng, Chenkai and Zhao, Xiaoyu and Yang, Yichun and Wu, Gangshan and Wang, Limin},
    title     = {SportsMOT: A Large Multi-Object Tracking Dataset in Multiple Sports Scenes},
    booktitle = ICCV,
    month     = {October},
    year      = {2023},
    pages     = {9921-9931}
}

@InProceedings{deliege_soccernet-v2_2021,
    author    = {Deliege, Adrien and Cioppa, Anthony and Giancola, Silvio and Seikavandi, Meisam J. and Dueholm, Jacob V. and Nasrollahi, Kamal and Ghanem, Bernard and Moeslund, Thomas B. and Van Droogenbroeck, Marc},
    title     = {SoccerNet-v2: A Dataset and Benchmarks for Holistic Understanding of Broadcast Soccer Videos},
    booktitle = CVPRW,
    month     = {June},
    year      = {2021},
    pages     = {4508-4519}
}

@InProceedings{wang_lvbench_2025,
    author    = {Wang, Weihan and He, Zehai and Hong, Wenyi and Cheng, Yean and Zhang, Xiaohan and others},
    title     = {LVBench: An Extreme Long Video Understanding Benchmark},
    booktitle = ICCV,
    month     = {October},
    year      = {2025},
    pages     = {22958-22967}
}

@article{lee_noah_2025,
  title={NOAH: Benchmarking Narrative Prior driven Hallucination and Omission in Video Large Language Models},
  author={Lee, Kyuho and Kim, Euntae and Choi, Jinwoo and Chang, Buru},
  journal={arXiv preprint arXiv:2511.06475},
  year={2025}
}

@article{lu_elv-halluc_2025,
  title={Elv-halluc: Benchmarking semantic aggregation hallucinations in long video understanding},
  author={Lu, Hao and Wang, Jiahao and Zhang, Yaolun and Wang, Ruohui and Zheng, Xuanyu and Tang, Yepeng and Lin, Dahua and Lu, Lewei},
  journal={arXiv preprint arXiv:2508.21496},
  year={2025}
}

@InProceedings{li_vidhalluc_nodate,
    author    = {Li, Chaoyu and Im, Eun Woo and Fazli, Pooyan},
    title     = {VidHalluc: Evaluating Temporal Hallucinations in Multimodal Large Language Models for Video Understanding},
    booktitle = CVPR,
    month     = {June},
    year      = {2025},
    pages     = {13723-13733}
}

@InProceedings{rawal_argus_2025,
    author    = {Rawal, Ruchit and Shirkavand, Reza and Huang, Heng and Somepalli, Gowthami and Goldstein, Tom},
    title     = {ARGUS: Hallucination and Omission Evaluation in Video-LLMs},
    booktitle = ICCV,
    month     = {October},
    year      = {2025},
    pages     = {20280-20290}
}

@article{yao2024minicpm,
  title={MiniCPM-V: A GPT-4V Level MLLM on Your Phone},
  author={Yao, Yuan and Yu, Tianyu and Zhang, Ao and Wang, Chongyi and Cui, Junbo and others},
  journal={arXiv preprint arXiv:2408.01800},
  year={2024}
}

@inproceedings{raoMatchTimeAutomaticSoccer2024a,
  title = {{{MatchTime}}: {{Towards}} Automatic Soccer Game Commentary Generation},
  shorttitle = {{{MatchTime}}},
  booktitle = {Proceedings of the 2024 {{Conference}} on {{Empirical Methods}} in {{Natural Language Processing}}},
  author = {Rao, Jiayuan and Wu, Haoning and Liu, Chang and Wang, Yanfeng and Xie, Weidi},
  year = 2024,
  month = nov,
  pages = {1671--1685},
  urldate = {2026-03-02}
}

@InProceedings{liDiscoveringSpatiotemporalRationales2023,
    author    = {Li, Yicong and Xiao, Junbin and Feng, Chun and Wang, Xiang and Chua, Tat-Seng},
    title     = {Discovering Spatio-Temporal Rationales for Video Question Answering},
    booktitle = ICCV,
    month     = {October},
    year      = {2023},
    pages     = {13869-13878}
}

@InProceedings{zhou_temporal_2018,
author = {Zhou, Bolei and Andonian, Alex and Oliva, Aude and Torralba, Antonio},
title = {Temporal Relational Reasoning in Videos},
booktitle = ECCV,
month = {September},
year = {2018}
}

@inproceedings{ren_testa_2023,
    title = "{TESTA}: Temporal-Spatial Token Aggregation for Long-form Video-Language Understanding",
    author = "Ren, Shuhuai  and
      Chen, Sishuo  and
      Li, Shicheng  and
      Sun, Xu  and
      Hou, Lu",
    booktitle = "Findings of the Association for Computational Linguistics: EMNLP 2023",
    month = dec,
    year = "2023",
    address = "Singapore",
    publisher = "Association for Computational Linguistics",
    pages = "932--947",
}

@article{li_temporal_2025,
  title={Temporal preference optimization for long-form video understanding},
  author={Li, Rui and Wang, Xiaohan and Zhang, Yuhui and Zohar, Orr and Wang, Zeyu and Yeung-Levy, Serena},
  journal={arXiv preprint arXiv:2501.13919},
  year={2025}
}

@article{wang_time-r1_2025,
  title={Time-r1: Post-training large vision language model for temporal video grounding},
  author={Wang, Ye and Wang, Ziheng and Xu, Boshen and Du, Yang and Lin, Kejun and Xiao, Zihan and Yue, Zihao and Ju, Jianzhong and Zhang, Liang and Yang, Dingyi and others},
  journal={arXiv preprint arXiv:2503.13377},
  year={2025}
}

@InProceedings{ye_re-thinking_2025,
    author    = {Ye, Jinhui and Wang, Zihan and Sun, Haosen and Chandrasegaran, Keshigeyan and Durante, Zane and others},
    title     = {Re-thinking Temporal Search for Long-Form Video Understanding},
    booktitle = CVPR,
    month     = {June},
    year      = {2025},
    pages     = {8579-8591}
}

@article{vteam2025glm45vglm41vthinkingversatilemultimodal,
  title={Glm-4.5 v and glm-4.1 v-thinking: Towards versatile multimodal reasoning with scalable reinforcement learning},
  author={Hong, Wenyi and Yu, Wenmeng and Gu, Xiaotao and Wang, Guo and Gan, Guobing and Tang, Haomiao and Cheng, Jiale and Qi, Ji and Ji, Junhui and Pan, Lihang and others},
  journal={arXiv preprint arXiv:2507.01006},
  year={2025}
}

@article{luOvis25TechnicalReport2025a,
  title={Ovis2. 5 technical report},
  author={Lu, Shiyin and Li, Yang and Xia, Yu and Hu, Yuwei and Zhao, Shanshan and Ma, Yanqing and Wei, Zhichao and Li, Yinglun and Duan, Lunhao and Zhao, Jianshan and others},
  journal={arXiv preprint arXiv:2508.11737},
  year={2025}
}

@article{anLLaVAOneVision15FullyOpen2025,
  title={Llava-onevision-1.5: Fully open framework for democratized multimodal training},
  author={An, Xiang and Xie, Yin and Yang, Kaicheng and Zhang, Wenkang and Zhao, Xiuwei and Cheng, Zheng and Wang, Yirui and Xu, Songcen and Chen, Changrui and Zhu, Didi and others},
  journal={arXiv preprint arXiv:2509.23661},
  year={2025}
}

@article{shaoDeepSeekMathPushingLimits2024c,
  title={Deepseekmath: Pushing the limits of mathematical reasoning in open language models},
  author={Shao, Zhihong and Wang, Peiyi and Zhu, Qihao and Xu, Runxin and Song, Junxiao and Bi, Xiao and Zhang, Haowei and Zhang, Mingchuan and Li, YK and Wu, Yang and others},
  journal={arXiv preprint arXiv:2402.03300},
  year={2024}
}

@inproceedings{kojimaLargeLanguageModels2023a,
 author = {Kojima, Takeshi and Gu, Shixiang (Shane) and Reid, Machel and Matsuo, Yutaka and Iwasawa, Yusuke},
 booktitle = NeurIPS,
 editor = {S. Koyejo and S. Mohamed and A. Agarwal and D. Belgrave and K. Cho and A. Oh},
 pages = {22199--22213},
 publisher = {Curran Associates, Inc.},
 title = {Large Language Models are Zero-Shot Reasoners},
 volume = {35},
 year = {2022}
}

@article{xuSlowFastLLaVA15FamilyTokenefficient2025,
  title={Slowfast-llava-1.5: A family of token-efficient video large language models for long-form video understanding},
  author={Xu, Mingze and Gao, Mingfei and Li, Shiyu and Lu, Jiasen and Gan, Zhe and Lai, Zhengfeng and Cao, Meng and Kang, Kai and Yang, Yinfei and Dehghan, Afshin},
  journal={arXiv preprint arXiv:2503.18943},
  year={2025}
}

@article{teamGLM45AgenticReasoning2025,
  title={Glm-4.5: Agentic, reasoning, and coding (arc) foundation models},
  author={Zeng, Aohan and Lv, Xin and Zheng, Qinkai and Hou, Zhenyu and Chen, Bin and Xie, Chengxing and Wang, Cunxiang and Yin, Da and Zeng, Hao and Zhang, Jiajie and others},
  journal={arXiv preprint arXiv:2508.06471},
  year={2025}
}

@misc{bai2025qwen25vltechnicalreport,
      title={Qwen2.5-VL Technical Report}, 
      author={Shuai Bai and Keqin Chen and Xuejing Liu and Jialin Wang and Wenbin Ge and Sibo Song and Kai Dang and Peng Wang and Shijie Wang and others},
      year={2025},
      eprint={2502.13923},
      archivePrefix={arXiv},
      primaryClass={cs.CV}
}

@misc{ding2026retrievingrelevantmomentsbenchmark,
      title={Retrieving Any Relevant Moments: Benchmark and Models for Generalized Moment Retrieval}, 
      author={Yiming Ding and Siyu Cao and Luyuan Jiao and Yixuan Li and Zitong Wang and Zhiyong Liu and Lu Zhang},
      year={2026},
      eprint={2605.02623},
      archivePrefix={arXiv},
      primaryClass={cs.CV},
      url={https://arxiv.org/abs/2605.02623}, 
}

\end{document}